%% file: neurips_2026.tex
\documentclass{article}

\newcommand{\ours}{GLINT}

    \PassOptionsToPackage{numbers, compress}{natbib}
 \usepackage[main, final]{neurips_2026}

\usepackage[utf8]{inputenc} 
\usepackage[T1]{fontenc}    
\usepackage{hyperref}       
\usepackage{url}            
\usepackage{booktabs}       
\usepackage{amsfonts}       
\usepackage{nicefrac}       
\usepackage{microtype}      
\usepackage{xcolor}         

\usepackage{array}
\usepackage{multirow}
\usepackage{makecell}
\usepackage{graphicx}
\usepackage{pifont}
\usepackage[table]{xcolor}
\usepackage{tikz}
\usepackage{nicematrix}
\usepackage{subcaption}
\usepackage{wrapfig}

\title{\ours: Sparsely Gated Vision-Language Alignment for Fine-Grained Radiology Representations}

\author{%
  \textbf{Jonggwon Park}$^{1*\dagger}$ \hskip0.9em
  \textbf{Seongeun Lee}$^{1*}$ \hskip0.9em
  \textbf{Junhyun Park}$^{1\ddagger}$ \hskip0.9em
  \textbf{Hannah Yun}$^{1\ddagger}$ \\
  \textbf{Hyunwoong Kim}$^{1}$ \hskip0.9em
  \textbf{Sohyun Jeong}$^{1}$ \hskip0.9em
  \textbf{Hyewon Kang}$^{1}$ \hskip0.9em
  \textbf{Byungmu Yoon}$^{1}$ \hskip0.9em
  \textbf{Kyoyun Choi}$^{2}$ \\
  \\
  $^{1}$DEEPNOID Inc., Seoul, South Korea \\
  $^{2}$Department of Artificial Intelligence and Data Science, Sejong University, Seoul, South Korea \\[4pt]
  {\tt jgpark@deepnoid.com, selee@deepnoid.com}
}
\begin{document}
\maketitle
\renewcommand{\thefootnote}{\fnsymbol{footnote}}
\footnotetext[1]{Equal contribution.}
\footnotetext[2]{Corresponding author.}
\footnotetext[3]{Equal second-author contribution.}
\renewcommand{\thefootnote}{\arabic{footnote}}
\setcounter{footnote}{0}

\input{sec/0_abstract}    
\input{sec/1_intro}
\input{sec/2_related_works}
\input{sec/3_methods}
\input{sec/4_experiments}

\input{sec/5_results}
\input{sec/6_conclusion}

{
\small
\bibliographystyle{plainnat}
\bibliography{custom}
}

\appendix

\input{sec/X_appendix}

\clearpage

\input{checklist.tex}

\end{document}

%% file: sec/0_abstract.tex
\begin{abstract}
Vision-language models (VLMs) for radiology have emerged as a scalable paradigm by leveraging image-report pairs naturally produced in clinical workflows.
However, this pairing reveals a mismatch in scale: each finding occupies only a small region of the image, yet supervision is provided only at the global image-report level.
This poses a central challenge: prior approaches spread weight densely across all patches rather than concentrating on the sparse subset relevant to a given query.
To address this, we present \textbf{\ours} (\textbf{G}ated \textbf{L}anguage-\textbf{I}mage alignme\textbf{NT}), a framework that models this sparse correspondence.
On the alignment side, we introduce \emph{Sparsely Gated Alignment}, a novel architecture in which a sigmoid gate over a separate gate embedding space activates only the patches relevant to each textual query, yielding sparsity.
On the representation side, we add \emph{Dense Feature Regularization}, which anchors the trainable encoder's intermediate features to a frozen self-supervised learning (SSL) teacher, preserving the fine-grained patch features that the gate relies on.
The same recipe applies to both 2D chest X-ray (CXR) and 3D chest computed tomography (CT), built with DINOv3 and V-JEPA 2.1, respectively.
\ours\ enables zero-shot classification, grounding, and segmentation from free-text queries, and to our knowledge is the first to demonstrate zero-shot segmentation on 3D CT volumes without mask supervision.
Notably, the most pronounced gains arise on zero-shot grounding and segmentation, where sparse, query-specific localization is required, consistent with our design intent.
In downstream evaluation, \ours\ outperforms both SSL encoders
and medical VLMs on classification, report generation, and segmentation. Code is available at \url{https://github.com/deepnoid-ai/GLINT}.
\end{abstract}

%% file: sec/1_intro.tex
\section{Introduction}
\label{sec:introduction}

Vision-language models (VLMs) have emerged as a scalable paradigm for medical image analysis~\citep{mgca, multimodal-med-gemini}, particularly in radiology, where each image is routinely paired with a free-text report~\citep{mimiccxr, ctclip}. Report-based supervision removes the need for manual annotation~\citep{litjens2017} and allows models to learn from a long tail of findings beyond a fixed label set~\citep{cxrlt2025}. It also mitigates the inter-annotator disagreement that limits pixel-level supervision~\citep{warfield2004staple}. 
These VLMs enable language-centric clinical applications such as report drafting~\citep{seah2025drafting}.

\begin{figure*}[t]                                     
    \centering
    \includegraphics[width=\textwidth]{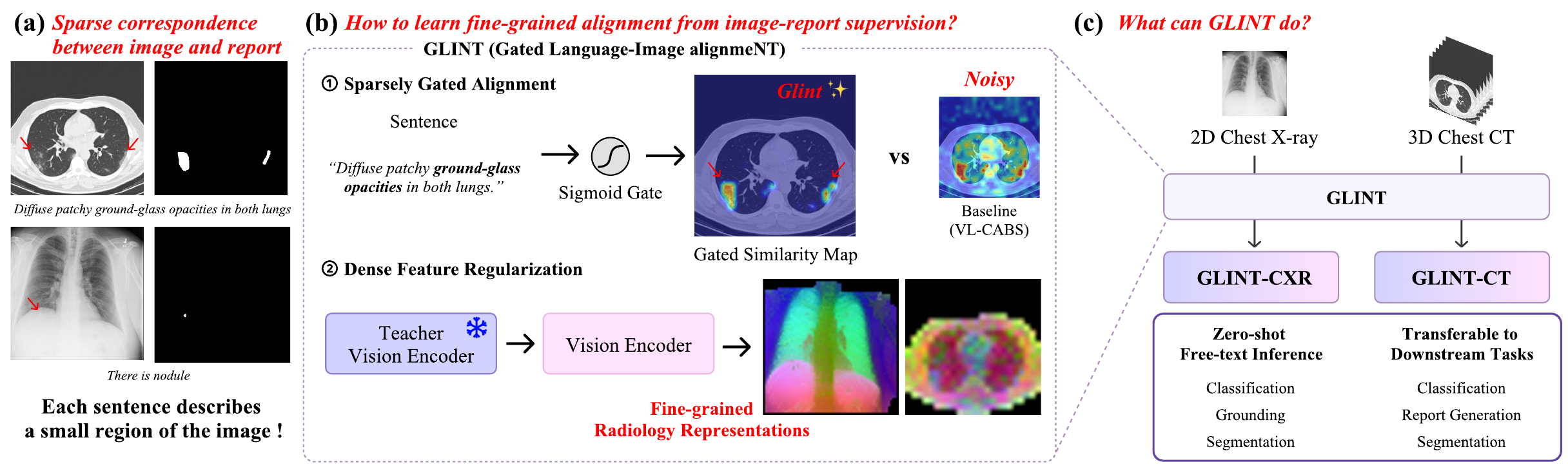} 
    \caption{
     Overview of \ours.
    \textbf{(a)} Each report sentence corresponds to a small region.
    \textbf{(b)} \ours\ combines \emph{Sparsely Gated Alignment} (sigmoid-gated patch weighting) with \emph{Dense Feature Regularization} (anchoring patches to a frozen SSL teacher).
    \textbf{(c)} The same recipe applies to 2D CXR and 3D CT.}
    \label{fig:intro}                                    
  \end{figure*}

Realizing this potential in radiology, however, faces a core challenge: a mismatch in scale.
Each finding occupies only a small region of the image, such as a sub-centimeter pulmonary nodule or a focal consolidation, paired with a specific span of text in the report~\citep{mscxr} (Figure~\ref{fig:intro}~(a)).
Yet supervision is provided only at the global image-report level, leaving the alignment between text and image regions implicit.
Across both 2D images and 3D volumes, recovering this fine-grained alignment from image-level supervision alone remains the central challenge for radiology VLMs~\citep{gloria, mgca}.

Prior approaches address this challenge along two axes, and each falls short. On the alignment side, methods rely on global image-report matching~\citep{ctclip, merlin}, cross-modal attention~\citep{gloria, mgca, carzero}, or softmax-normalized patch aggregation~\citep{radzero};  these spread alignment weight across all patches rather than concentrating it on the sparse subset relevant to a given report. On the representation side, self-supervised learning (SSL) foundations~\citep{dinov3, vjepa21} now provide strong patch-level features for medical imaging~\citep{raddino, chexworld}, but preserving them during language adaptation is non-trivial: fine-tuning distorts pretrained features~\citep{kumar2022finetuning}, while freezing the image encoder keeps them intact~\citep{lit} but limits cross-modal alignment.
These limitations also manifest unevenly across modalities: while zero-shot localization from free-text queries is well established on chest X-ray (CXR)~\citep{mscxr, radzero}, on 3D chest computed tomography (CT) it remains largely unexplored~\citep{ctclip, merlin, radzero3d}.

To address these limitations, we present \textbf{\ours}~(\textbf{G}ated \textbf{L}anguage-\textbf{I}mage alignme\textbf{NT}), a framework that models this sparse, fine-grained alignment~(Figure~\ref{fig:intro}). 
For alignment, we introduce Sparsely Gated Alignment (SGA), a novel architecture in which patch and text features are projected into a separate gate embedding space, where a sigmoid gate learns to sparsely activate the patches relevant to each text query. 
For representation, we add \emph{Dense Feature Regularization} (DFR), which anchors the encoder's
intermediate features to a frozen SSL teacher, preserving fine-grained patch representations during VL training. The same recipe applies to both 2D CXR and 3D CT, built on DINOv3~\citep{dinov3} and V-JEPA 2.1~\citep{vjepa21}, respectively. These design choices give
\ours\ two distinguishing strengths. \textbf{Zero-shot free-text inference}: \ours\ supports zero-shot classification, grounding, and segmentation from free-text queries. To our knowledge, it is the first to demonstrate zero-shot segmentation on 3D CT volumes without mask supervision.
\textbf{Fine-grained radiology representations}: \ours\ outperforms SSL encoders and medical VLMs on classification, report generation, and supervised segmentation across CXR and CT.

%% file: sec/2_related_works.tex
\section{Related Works} \label{sec:related_works}

\subsection{Vision-Language Alignment in Radiology}

For chest X-ray, ConVIRT~\cite{convirt} and MedCLIP~\cite{medclip} perform global image-report alignment, while GLoRIA~\cite{gloria} aligns patches with words, MGCA~\cite{mgca} extends this to multiple granularities, and MAVL~\cite{mavl} decomposes alignment by disease. MedKLIP~\cite{medklip} extracts structured entities, and KAD~\cite{kad} injects medical knowledge graphs into training. 
RadZero~\cite{radzero} introduces VL-CABS, a softmax-normalized patch aggregation architecture that enables zero-shot multi-task inference on CXR. 
For 3D CT, CT-CLIP~\cite{ctclip} adapts CLIP~\cite{clip} via volume-report alignment, and Merlin~\cite{merlin} combines EHR phenotype supervision with report-level contrastive learning.
fVLM~\cite{fvlm} and ViSD-Boost~\cite{visd-boost} enable fine-grained alignment by decomposing volumes into anatomy-specific components, both relying on TotalSegmentator~\cite{totalsegmentator} for anatomical priors. RadZero3D~\cite{radzero3d} adapts a video SSL model to chest CT, and COLIPRI~\cite{colipri} unifies masked image modeling, report generation, and contrastive learning under a multi-task framework.

Only a small subset of these methods further extends to zero-shot localization (grounding or segmentation from text without localization supervision). For chest X-ray, MedKLIP~\cite{medklip} and CARZero~\cite{carzero} derive localization from cross-attention
maps, 
while RadZero~\cite{radzero} derives localization from patch-text similarities.
For 3D CT, however, zero-shot localization remains largely unexplored: 
RadZero3D extends this approach to chest CT, but as noted in its analysis, voxel-level localization remains limited,
and prior 3D VLMs~\cite{ctclip, fvlm, visd-boost} primarily target zero-shot classification. Alternatively, VoxTell~\cite{voxtell} supports free-text prompts but requires large-scale voxel-level mask supervision. In contrast, \ours\ adopts a sparsely gated alignment that concentrates alignment onto a small subset of relevant patches, enabling, to our knowledge, the first zero-shot free-text segmentation on 3D CT without any mask supervision.

\subsection{SSL Representations in Vision-Language Models}
Self-supervised foundation models~\cite{dinov2, dinov3, vjepa2, vjepa21} produce strong fine-grained patch-level representations that have been widely adopted in medical imaging~\cite{raddino, chexworld, 3dino}. However, fine-tuning distorts pretrained features~\cite{kumar2022finetuning}, and frozen encoders often outperform fine-tuning under global contrastive objectives~\cite{lit}, motivating work on retaining SSL features
during VL adaptation. TIPS~\cite{tips} and SigLIP~2~\cite{siglip2} integrate SSL objectives with contrastive alignment during pretraining, and VIRAL~\cite{viral} regularizes multimodal LLMs by aligning their internal visual representations with those of vision foundation models. For radiology, RadZero~\cite{radzero} keeps its SSL encoder frozen, RAD-DINO~\cite{raddino} continues SSL pretraining on large-scale CXR data without language supervision, and MRM~\cite{mrm} and COLIPRI~\cite{colipri} retain dense features by adding masked image modeling alongside language objectives. Rather than freezing the encoder or relying on continual SSL pretraining alone, \ours\ adapts the encoder while anchoring its
intermediate features to a frozen SSL teacher, retaining fine-grained representations during training.

%% file: sec/3_methods.tex
\section{Methods}
\label{sec:methods}

\begin{figure*}[t]
  \centering
  \includegraphics[width=\textwidth]{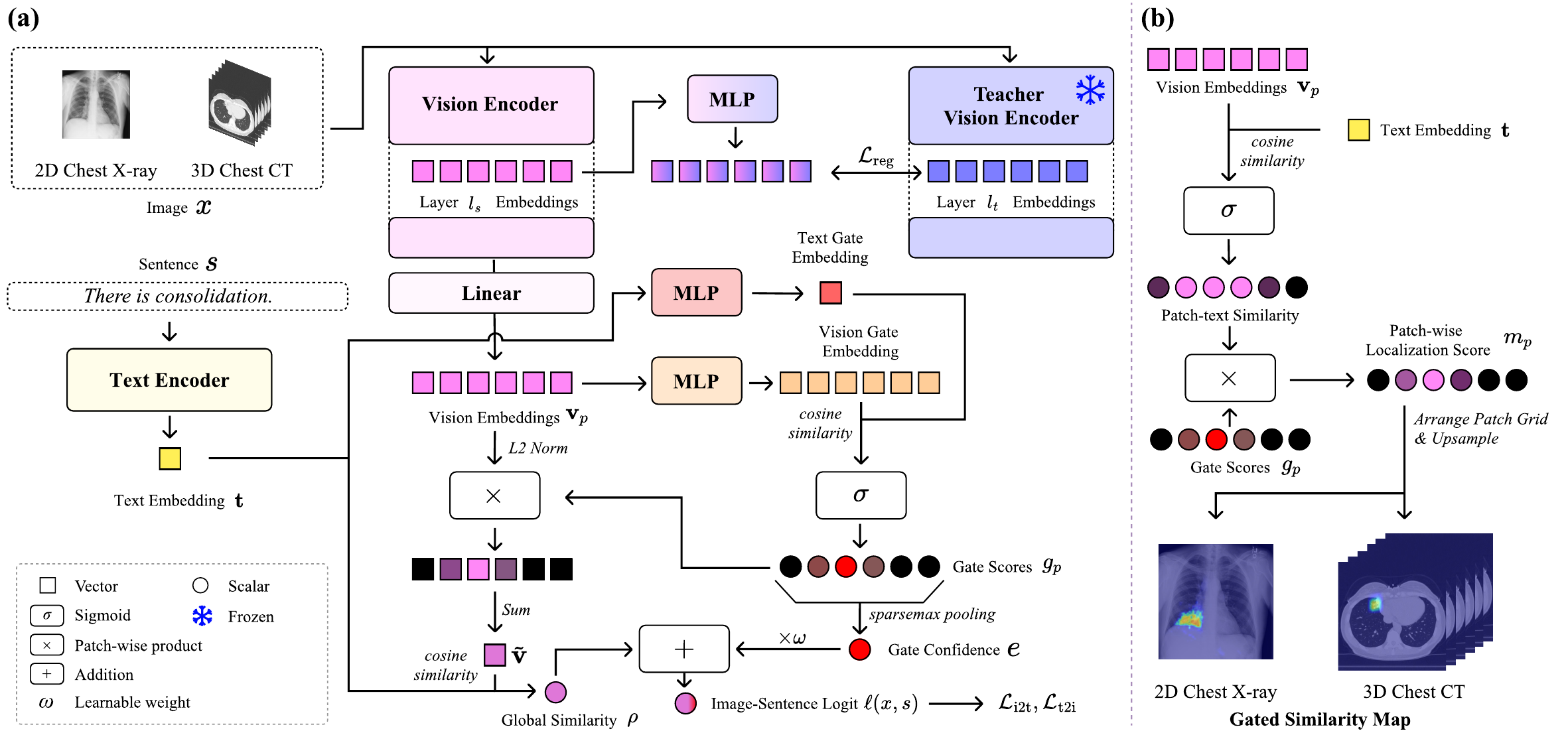}
    \caption{
    The overall framework of \ours.
    \textbf{(a)} model architecture, jointly showing SGA and DFR modules.
    \textbf{(b)} inference procedure for the gated similarity map.
  }
  \label{fig:method}
\end{figure*}

\ours\ consists of two components: Sparsely Gated Alignment (Section~\ref{sec:gated-alignment}) and Dense Feature Regularization  
(Section~\ref{sec:ssl-distillation}). Figure~\ref{fig:method} summarizes the overall design.

\subsection{Text Supervision}
\label{sec:text_supervision}

Following prior work~\cite{carzero, radzero}, we treat each radiology report as a set of independent observations rather than a single global caption.
Specifically, an LLM~\cite{agarwal2025gpt} decomposes the report into sentences, each describing a single observation; we detail the prompt design in Appendix~\ref{sec:appendix_prompts}.

Given a report decomposed into $N$ sentences
$\{s_n\}_{n=1}^{N}$, we obtain sentence embeddings
$\mathbf{t}_n = f_t(s_n) \in \mathbb{R}^{d}$
via the text encoder $f_t$, where $d$ is the shared
image-text embedding dimension.
We use these as alignment targets paired with the
corresponding image during training.

\subsection{Sparsely Gated Alignment}
\label{sec:gated-alignment}

Given an input image $x$ (a 2D chest X-ray or a 3D CT volume), a modality-specific vision encoder $f_v$ produces $P$ patch-level features on a 2D or 3D patch grid, which we project to the shared dimension $d$ via a linear layer, yielding $\{\mathbf{v}_p\}_{p=1}^{P} \subset \mathbb{R}^{d}$, where $p$ indexes patches.
We write $\mathbf{t} = f_t(s)$ for the embedding of a single sentence $s$, dropping the index $n$ for brevity.
Most prior alignment architectures aggregate the patch embeddings into a sentence-conditioned vision representation either via cross-modal attention~\cite{gloria, mgca, carzero} or via softmax-normalized patch aggregation~\cite{radzero}, spreading alignment weight across many patches rather than concentrating it on the relevant region.
We instead introduce \emph{Sparsely Gated Alignment} (SGA), inspired by recent work on gated attention~\cite{qiu2026gated}:
an architecture that activates the relevant subset of patches via a learnable gate.

\textbf{Gate scores.}
On top of the patch and text embeddings, we apply a gate
head with separate two-layer MLP branches, $\phi_t$ for
text and $\phi_v$ for vision, that produce $d$-dimensional gate embeddings.
Patch-wise gate scores are then obtained by applying a
sigmoid to the temperature-scaled cosine similarity between
text and vision gate embeddings, 
\begin{equation}
g_p = \sigma\left(
    \cos\bigl(\phi_t(\mathbf{t}),\,\phi_v(\mathbf{v}_p)\bigr)
    / \tau_g
\right) \in (0,1),
\label{eq:gate_scores}
\end{equation}
where $\sigma$ is the sigmoid function, $\cos$ denotes
cosine similarity, and $\tau_g$ is a learnable temperature
for the gate.
Unlike softmax, the sigmoid does not normalize across patches: the gate vector $\mathbf{g} = (g_1,\ldots,g_P)^\top$ can be sparsely active (a few patches relevant, the rest irrelevant) or uniformly small (no spatial referent).
This relaxes the implicit one-hot prior of softmax aggregation and lets the gate adapt to the actual support of each sentence.

\textbf{Gated aggregation.}
The aggregated vision representation $\tilde{\mathbf{v}}$ for sentence $s$ is the gated sum of L2-normalized patch embeddings, so that each patch's contribution depends only on its gate score:
\begin{equation}
\tilde{\mathbf{v}} = \sum_{p=1}^{P} g_p \frac{\mathbf{v}_p}{\|\mathbf{v}_p\|_2}.
\end{equation}
The cosine similarity between the text embedding and the aggregated vision representation is
\begin{equation}
\rho = \cos\bigl(\mathbf{t},\, \tilde{\mathbf{v}}\bigr).
\end{equation}

\textbf{Gate confidence shift.}
Cosine similarity discards the magnitude of $\tilde{\mathbf{v}}$, so a sentence with no spatial referent (gate $\approx 0$) cannot be distinguished from one whose aggregated vision representation happens to align in direction.
To restore this signal, we add a logit shift driven by the gate's overall confidence.
We pool $\mathbf{g}$ with sparsemax weights
$\mathbf{w} = \mathrm{sparsemax}(\mathbf{g})$~\cite{martins2016sparsemax}
to obtain a scalar gate confidence
$e = \mathbf{w}^{\top}\mathbf{g}$.

Sparsemax can produce a sparse distribution that concentrates on the highest gate values, so $e$ behaves as a smoothed top-$k$ pool of gate confidence.
This differs from both mean pooling, which dilutes the signal across irrelevant patches, and max pooling, which is sensitive to single-patch noise. The final image-sentence logit is $\ell(x, s) = \rho + \omega \cdot e$, where $\omega$ is a learnable weight. The shift makes $\ell$ sensitive to absolute gate activation, allowing the model to express absent versus present findings without violating the sparsity-inducing role of the gate.

\subsection{Dense Feature Regularization}
\label{sec:ssl-distillation}

Training the encoder for VL alignment with an image-level contrastive loss provides no direct supervision over patch features. Such objectives are known to cause representation drift in pretrained features~\cite{kumar2022finetuning}, while freezing the encoder avoids drift at the cost of adaptation~\cite{lit}. SGA needs both: language adaptation and intact patch features for discriminative gating. We therefore add \emph{Dense Feature Regularization} (DFR), which anchors the trainable student's patch representations to those of a frozen SSL teacher, in a similar spirit to RLHF's KL penalty toward a reference policy~\cite{ouyang2022training}.

DFR follows intermediate-layer alignment~\cite{viral, ypsilantis2025infusing},
using a frozen teacher from the same SSL family as the student $f_v$.
Both encoders share the same patch grid.
For a mini-batch of $B$ images, we extract patch token embeddings
$\mathbf{H}_s \in \mathbb{R}^{B \times P \times d_s}$
from layer $l_s$ of the student and
$\mathbf{H}_t \in \mathbb{R}^{B \times P \times d_t}$
from layer $l_t$ of the teacher, where $b$ indexes images and $d_s$, $d_t$ are the
student and teacher feature dimensions, respectively.
A multi-layer perceptron (MLP) projector
$h: \mathbb{R}^{d_s} \rightarrow \mathbb{R}^{d_t}$
maps student features to the teacher's dimension.
The loss
$\mathcal{L}_{\text{reg}}$ uses the mean per-patch cosine
distance between projected student features and teacher features:
\begin{equation}
\mathcal{L}_{\text{reg}} = \frac{1}{BP} \sum_{b=1}^{B} \sum_{p=1}^{P} \bigl(1 - \cos(h(\mathbf{H}_s[b, p]), \mathbf{H}_t[b, p])\bigr).
\end{equation}

The regularizer anchors the student to the teacher at the patch level, preserving fine-grained pretrained features while the encoder adapts to VL alignment.

\subsection{Training Objective}
\label{sec:training-objective}

We train with bidirectional contrastive losses on the image-sentence
logits $\ell(x,s)$, following CLIP image-text alignment~\cite{clip}.
Consider a mini-batch of $N_s$ sentences $\{s_n\}_{n=1}^{N_s}$ drawn from
$N_i$ images $\{x_m\}_{m=1}^{N_i}$, where $n$ indexes sentences,
$m$ indexes images, and $\pi(n)$ is the index of the image whose
report contains $s_n$.
Let $\ell_{m,n} = \ell(x_m,s_n)/\tau_l$ be the temperature-scaled
logit, with $\tau_l$ a learnable loss temperature.

The text-to-image branch is standard InfoNCE, with the associated
image $x_{\pi(n)}$ as the positive and all other images in the batch
as negatives:
\begin{equation}
\mathcal{L}_{\text{t2i}} =
-\frac{1}{N_s}\sum_{n=1}^{N_s}
\log\frac{\exp(\ell_{\pi(n),n})}
{\sum_{m=1}^{N_i}\exp(\ell_{m,n})}.
\label{eq:t2i}
\end{equation}

The image-to-text branch uses the multi-positive contrastive
(MP-NCE) loss~\cite{uniclip, radzero}.
For each positive pair, the denominator includes its positive
term and negative terms from sentences $s_k$ paired with other
images; other sentences from the same report are excluded
from the negatives:
\begin{equation}
\mathcal{L}_{\text{i2t}} =
-\frac{1}{N_s}\sum_{n=1}^{N_s}
\log\frac{\exp(\ell_{\pi(n),n})}
{\exp(\ell_{\pi(n),n}) +
 \sum_{k:\,\pi(k)\ne \pi(n)}\exp(\ell_{\pi(n),k})}.
\label{eq:i2t}
\end{equation}

The alignment loss is
$\mathcal{L}_{\text{align}} =
(\mathcal{L}_{\text{t2i}} + \mathcal{L}_{\text{i2t}})/2$,
and the full training loss combines it with the DFR loss as
$\mathcal{L} = \mathcal{L}_{\text{align}} +
\lambda\,\mathcal{L}_{\text{reg}}$,
where $\lambda$ is a fixed weight balancing the regularization.
\subsection{Inference}
\label{sec:inference}

\textbf{Zero-shot classification.} Given an image and a textual prompt, we compute the classification
score $\sigma(\ell(x,s)/\tau_l)$, using the same temperature
$\tau_l$ learned during training.

\textbf{Zero-shot localization.} For grounding and segmentation, we compute a patch-wise localization score $m_p = g_p \cdot \sigma\bigl(\cos(\mathbf{t}, \mathbf{v}_p) / \tau_l\bigr)$ for each patch $p$, combining the gate with patch-text similarity.
The gate $g_p$ captures the spatial selectivity learned during alignment, while the sigmoid-scaled cosine measures direct patch-text similarity in the embedding space; their product is high only at patches where both signals agree.
We arrange $\{m_p\}$ on the patch grid and upsample to the input image resolution, bilinearly for 2D CXR and trilinearly for 3D CT. The resulting \emph{Gated Similarity Map} (GSM) is used for both grounding and segmentation.

%% file: sec/4_experiments.tex
\section{Experiments}
\label{sec:experiments}

\subsection{Training Datasets}
\label{sec:training_datasets}

\textbf{Chest X-ray.}
MIMIC-CXR v2.0.0~\cite{mimiccxr} contains 227,827 radiology reports describing 377,110 CXR images from 65,379 patients.
Following the official split, we retain only studies from which decomposed sentences (described in Section~\ref{sec:text_supervision}) can be extracted, resulting in 353,413 training images and 2,853 validation images.

\textbf{Chest CT.}
CT-RATE v2~\cite{ctclip} comprises 50,188 chest CT volumes with paired radiology reports and multi-abnormality labels.
We use the official validation split (3,002 volumes) as the test set,
and construct a validation set by randomly sampling patients from
the official training split.
All three sets are patient-disjoint.
Following the same report processing and the exclusions described
in Appendix~\ref{apx:test_dataset}, the final data for vision-language
alignment consist of 43,337 training volumes and 2,990 validation volumes.

\subsection{Evaluation Datasets}
\label{sec:evaluation_datasets}

\textbf{Chest X-ray zero-shot evaluation.}
Classification is evaluated on PadChest~\cite{padchest} (192 classes) and its PadChest20~\cite{carzero} subset (20 rare classes), ChestXray14~\cite{chestxray14} (14 classes) and its ChestX-Det10~\cite{chestXdet10} subset (10 classes), CheXpert~\cite{chexpert} (5 classes), and Open-I~\cite{openi} (18 classes).
Visual grounding is evaluated on ChestX-Det10 with bounding boxes for 10 disease classes, and MS-CXR~\cite{mscxr} with bounding boxes paired with free-text phrases from radiology reports.
Segmentation is evaluated on positive cases from ChestX-Det~\cite{chestXdet} (13 classes) and SIIM~\cite{siim} (pneumothorax), both with pixel-level annotations.

\textbf{Chest X-ray downstream evaluation.}
For downstream evaluation, classification is evaluated on four CXR datasets: ChestXray14, VinDr-CXR~\cite{vindrcxr} (27 classes), RSNA Pneumonia (RSNA-PN)~\cite{rsna}, and SIIM. 
Segmentation is evaluated on ChestX-Det~\cite{chestXdet} (13 classes) and SIIM.
Report generation is evaluated on the Findings sections of MIMIC-CXR reports, with the decoder trained on a random 10\% subset of the training set.

\textbf{Chest CT zero-shot evaluation.}
We perform zero-shot classification on CT-RATE~\cite{ctclip} (18 classes) for internal evaluation, and on RAD-ChestCT~\cite{radchestct} (16 classes) for external evaluation, following \citet{ctclip}.
Free-text segmentation is evaluated on ReXGroundingCT~\cite{rexgroundingct}, with voxel-level masks paired with free-text findings from radiology reports.

\textbf{Chest CT downstream evaluation.}
We evaluate on three downstream tasks. Classification is evaluated on CT-RATE, RSNA Pulmonary Embolism (RSNA-PE)~\cite{rsna_pe}, and LIDC-IDRI~\cite{lidc};
report generation on CT-RATE;
and segmentation on Task 6 (primary lung cancers) of the Medical Segmentation Decathlon (MSD-Lung)~\cite{antonelli2022medical} and the NSCLC-Radiomics dataset~\cite{aerts2014decoding}.
For NSCLC-Radiomics, we evaluate on the primary gross tumor volume (GTV).
Dataset details, splits, and evaluation protocols are provided in Appendix~\ref{apx:test_dataset}.

\subsection{Evaluation Metrics}
For multi-class tasks, we report macro-averaged metrics throughout.
For \textbf{classification}, area under the ROC curve (AUROC) is used.
\textbf{Grounding} performance is evaluated using the pointing-game protocol~\cite{pointinggame}, which measures whether the pixel with the maximum predicted value lies within the ground-truth bounding box.
For \textbf{segmentation}, we use the Dice score, following prior zero-shot~\cite{radzero} and downstream~\cite{chexworld,mgca} evaluation protocols.
\textbf{Report generation} is evaluated using both natural language generation (NLG) metrics and clinical efficacy metrics.
NLG metrics include ROUGE-L~\cite{rouge} and METEOR~\cite{meteor}.
For CXR, clinical efficacy is evaluated using the macro-averaged F1 score from CheXbert~\cite{chexbert} (14 classes) and F1-RadGraph~\cite{f1_radgraph} (using $\text{RG}_{\text{ER}}$); for CT, we use the macro-averaged F1 from a RadBERT-based labeler~\cite{ctclip} (18 classes) and CRG~\cite{crg}.
We report the mean $\pm$ standard deviation over three random seeds in Tables~\ref{tab:appendix_cxr_zeroshot_stat_sig}--\ref{tab:appendix_ct_downstream_stat_sig}.

\input{tables/cxr_zeroshot}
\input{tables/cxr_downstream}

\subsection{Implementation Details}
\label{sec:implementation_details}

For CXR, we use DINOv3~\cite{dinov3} with ViT-L (student) and ViT-7B (teacher) at $1024^2$ resolution. For chest CT, we adopt V-JEPA 2.1~\cite{vjepa21} with ViT-L (student) and ViT-G (teacher) at $384^2$.
CT volumes are preprocessed following \citet{ctclip} with MONAI~\cite{monai}, yielding $(D,H,W)=(240,384,384)$ inputs normalized to match V-JEPA 2.1's input distribution. The text encoder is initialized with MPNet (``all-mpnet-base-v2'')~\cite{sentence-bert}, and ``gpt-oss-120b''~\cite{agarwal2025gpt} performs sentence-level report decomposition. We set $d=768$ for both CXR and CT. The loss temperature is initialized at $\tau_l=0.07$,
following CLIP~\cite{clip}.
The gate temperature is initialized at $\tau_g=0.1$,
corresponding to the initial multiplicative logit scale
of $10$ used in SigLIP~\cite{siglip}.
Both temperatures are learnable and parameterized in log-space. The gate-confidence-shift scalar $\omega$ is initialized to 0, so the image-sentence logit starts without a shift and the shift gradually engages during training. For DFR, intermediate representations are extracted from the student layer $l_s=22$ and the teacher layer $l_t=36$ for CXR (out of 24 and 40 layers, respectively), and from $l_s=22$ and $l_t=44$ for chest CT (out of 24 and 48 layers, respectively), with fixed $\lambda=8$.  For all downstream tasks, the vision encoder remains frozen and only task-specific heads are trained. Please refer to Appendix~\ref{apx:implementation_detail} for downstream task details.

%% file: tables/cxr_zeroshot.tex
\begin{table*}[!t]
  \caption{Zero-shot performance on chest X-ray. Best results are highlighted in bold.}
  \centering
  \small
  \resizebox{\textwidth}{!}{
    \begin{NiceTabular}{l *{6}{wc{1.4cm}} *{2}{wc{1.6cm}} *{2}{wc{1.6cm}}}[color-inside]
      \toprule

      \Block[c]{3-1}{Method}
        & \multicolumn{6}{c}{\textbf{Classification}}
        & \multicolumn{2}{c}{\textbf{Grounding}}
        & \multicolumn{2}{c}{\textbf{Segmentation}} \\
      \specialrule{0pt}{2pt}{3pt}

        & PadChest & PadChest20 & ChestXray14 & ChestX-Det10 & CheXpert & Open-I
        & ChestX-Det10 & MS-CXR
        & ChestX-Det & SIIM \\
      \specialrule{0pt}{2pt}{3pt}

      & \multicolumn{6}{c}{Macro AUROC}
      & Macro Acc.
      & Acc.
      & Macro Dice
      & Dice \\
      \specialrule{0pt}{2pt}{0pt}
      \toprule

      GLoRIA$^\dagger$~\cite{gloria}
        & 0.565 & 0.558 & 0.610 & 0.645 & 0.750 & 0.589
        & 0.367 &   --
        &   --   &   --   \\

      KAD$^\dagger$~\cite{kad}
        & 0.750 & 0.735 & 0.789 & 0.735 & 0.905 & 0.807
        & 0.391 &   --
        &   --   &   --   \\

      MedKLIP$^\dagger$~\cite{medklip}
        & 0.629 & 0.688 & 0.726 & 0.713 & 0.879 & 0.759
        & 0.481 & 0.407
        & 0.249 & 0.044 \\

      CARZero$^\dagger$~\cite{carzero}
        & 0.810 & 0.837 & 0.811 & 0.796 & \textbf{0.923} & 0.838
        & 0.543 & 0.749
        & 0.304 & 0.081 \\

      MedSigLIP~\cite{medgemma}
        & 0.736 & 0.747 & 0.742 & 0.702 & 0.859 & 0.763
        &   --   &   --
        &   --   &   --   \\

      RadZero$^\dagger$~\cite{radzero}
        & 0.841 & 0.871 & 0.804 & 0.787 & 0.900 & 0.847
        & 0.622 & 0.844
        & 0.332 & 0.171 \\

      \specialrule{0pt}{0pt}{1pt}
      \rowcolor{gray!20}
      \ours-CXR
        & \textbf{0.863} & \textbf{0.917} & \textbf{0.825} & \textbf{0.853} & 0.921 & \textbf{0.881}
        & \textbf{0.729} & \textbf{0.892}
        & \textbf{0.434} & \textbf{0.317} \\
      \specialrule{0.7pt}{0pt}{-1pt}

      \CodeAfter
      \begin{tikzpicture}
          \begin{scope}[shorten <=2pt, shorten >=2pt]
            \draw (1-|2) -- (4-|2);
            \draw (1-|8) -- (4-|8);
            \draw (1-|10) -- (4-|10);
            \draw (4-|2) -- (11-|2);
            \draw (4-|8) -- (11-|8);
            \draw (4-|10) -- (11-|10);
        \end{scope}
        \begin{scope}[shorten <=3pt, shorten >=3pt]
            \draw (2-|2) -- (2-|8);   
            \draw (2-|8) -- (2-|10);  
            \draw (2-|10) -- (2-|12); 
            \draw (3-|2) -- (3-|8);
            \draw (3-|8) -- (3-|9);
            \draw (3-|9) -- (3-|10);
            \draw (3-|10) -- (3-|11);
            \draw (3-|11) -- (3-|12);
        \end{scope}
      \end{tikzpicture}

    \end{NiceTabular}
  }
    \label{tab:chest_xray_zeroshot}
\end{table*}

%% file: tables/cxr_downstream.tex
\begin{table*}[!t]
  \caption{
  Downstream task performance on chest X-ray,
  averaged over three random seeds.
  Bold denotes the highest mean performance.
  }
  \centering
  \small
  \setlength{\tabcolsep}{5pt}
  \renewcommand{\arraystretch}{1.15}

  \resizebox{\textwidth}{!}{
    \begin{NiceTabular}{l *{4}{wc{1.6cm}} *{4}{wc{1.6cm}} *{2}{wc{1.6cm}}}[color-inside]
      \toprule
      \Block[c]{3-1}{Method}

      & \multicolumn{4}{c}{\textbf{Classification}}
      & \multicolumn{4}{c}{\textbf{Report Generation}}
      & \multicolumn{2}{c}{\textbf{Segmentation}} \\
      \specialrule{0pt}{2pt}{3pt}

      & ChestXray14 & VinDr-CXR & RSNA-PN & SIIM
      & \multicolumn{4}{c}{MIMIC-CXR}
      & ChestX-Det & SIIM \\
      \specialrule{0pt}{2pt}{3pt}

      & \multicolumn{2}{c}{Macro AUROC}
      & \multicolumn{2}{c}{AUROC}
      & METEOR & ROUGE-L & CheXbert F1 & F1-RadGraph
      & Macro Dice & Dice \\
      \specialrule{0pt}{2pt}{0pt}
      \toprule

      DINOv3~\cite{dinov3}
      & 0.806 & 0.883 & 0.879 & 0.931 & 0.130 & 0.238 & 0.238 & 0.202 & 0.765 & 0.479 \\

      MGCA~\cite{mgca}
      & 0.808 & 0.923 & 0.892 & 0.928 & 0.131 & 0.237 & 0.259 & 0.204 & 0.779 & 0.514 \\

      KAD~\cite{kad}
      & 0.812 & 0.934 & 0.896 & 0.937 & 0.135 & 0.243 & 0.277 & 0.215 & 0.789 & 0.539 \\

      MedKLIP~\cite{medklip}
      & 0.801 & 0.919 & 0.893 & 0.937 & 0.135 & 0.237 & 0.268 & 0.204 & 0.781 & 0.507 \\

      MRM~\cite{mrm}
      & 0.818 & 0.907 & 0.893 & 0.937 & 0.133 & 0.240 & 0.271 & 0.209 & 0.793 & 0.577 \\

      MAVL~\cite{mavl}
      & 0.802 & 0.914 & 0.892 & 0.944 & 0.133 & 0.240 & 0.278 & 0.211 & 0.765 & 0.551 \\

      CheXWorld~\cite{chexworld}
      & 0.814 & 0.922 & 0.888 & 0.937 & 0.133 & 0.238 & 0.259 & 0.208 & 0.795 & 0.586 \\

      RAD-DINO~\cite{raddino}
      & 0.831 & 0.936 & 0.891 & 0.953 & 0.137 & 0.242 & 0.283 & 0.213 & 0.808 & 0.631 \\

      \specialrule{0pt}{0pt}{1pt}
      \rowcolor{gray!20}
      \ours-CXR
      & \textbf{0.847} & \textbf{0.944} & \textbf{0.902} & \textbf{0.975} & \textbf{0.143} & \textbf{0.249} & \textbf{0.314} & \textbf{0.225} & \textbf{0.818} & \textbf{0.730} \\
      \specialrule{0.7pt}{0pt}{-1pt}

      \CodeAfter
      \begin{tikzpicture}
          \begin{scope}[shorten <=2pt, shorten >=2pt]
            \draw (1-|2) -- (4-|2);
            \draw (1-|6) -- (4-|6);
            \draw (1-|10) -- (4-|10);
            \draw (4-|2) -- (13-|2);
            \draw (4-|6) -- (13-|6);
            \draw (4-|10) -- (13-|10);
        \end{scope}
        \begin{scope}[shorten <=3pt, shorten >=3pt]
            \draw (2-|2) -- (2-|6);   
            \draw (2-|6) -- (2-|10);  
            \draw (2-|10) -- (2-|12); 
            \draw (3-|2) -- (3-|4);
            \draw (3-|4) -- (3-|6);
            \draw (3-|6) -- (3-|10);
            \draw (3-|10) -- (3-|11);
            \draw (3-|11) -- (3-|12);

        \end{scope}
      \end{tikzpicture}
    \end{NiceTabular}
  } 
  \label{tab:cxr_downstream}
\end{table*}

%% file: sec/5_results.tex
\section{Results}
\label{sec:results}

\subsection{Zero-shot and Downstream Evaluation}
\label{sec:main_results}

\textbf{Chest X-ray zero-shot evaluation.}
We compare \ours-CXR against prior medical VLMs~\cite{gloria, kad, medklip, carzero, medgemma, radzero} (Table~\ref{tab:chest_xray_zeroshot}).
Results marked $\dagger$ are taken from RadZero~\cite{radzero},
while MedSigLIP and \ours-CXR are evaluated by us.
For GLoRIA and KAD, dashes indicate results not reported in the source.
MedSigLIP does not natively provide patch-text scores
for zero-shot localization.
\ours-CXR achieves the best or comparable performance across all benchmarks, with the most pronounced gains on dense localization over the prior state of the art: Macro Dice on multi-class ChestX-Det rises from 0.332 to 0.434, and SIIM pneumothorax Dice from 0.171 to 0.317. These gains follow from SGA's sparse, query-specific alignment, which concentrates on focal regions critical for grounding and segmentation.

\textbf{Chest X-ray downstream evaluation.}
Baselines span CXR VLMs~\cite{mgca, kad, medklip, mrm, mavl} and SSL methods~\cite{chexworld, raddino} (Table~\ref{tab:cxr_downstream}). \ours-CXR achieves the best performance on every downstream task, across classification, report generation, and segmentation. The largest gain appears on SIIM segmentation: our DINOv3~\cite{dinov3} initialization alone reaches only 0.479; vision-language alignment raises this to 0.730 Dice, surpassing RAD-DINO (0.631), an SSL foundation model continually trained on large-scale CXR data. This shows that DFR anchors \ours-CXR to its DINOv3 initialization, letting vision-language alignment build on rather than erase the pretrained features.

\input{tables/ct_zeroshot}
\input{tables/ct_downstream}

\textbf{Chest CT zero-shot evaluation.}
We compare \ours-CT against prior CT VLMs~\cite{ctclip, merlin, fvlm, visd-boost, radzero3d, colipri} on classification (Table~\ref{tab:ct_zeroshot}). \ours-CT achieves the best performance on both internal CT-RATE (0.850) and external RAD-ChestCT (0.789). 
The baselines marked $\dagger$ in Table~\ref{tab:ct_zeroshot} are taken from~\cite{visd-boost}, which reports under a protocol that excludes ``lymphadenopathy'' and ``medical material'' labels. Under the same protocol, \ours-CT achieves 0.855 Macro AUROC on CT-RATE and 0.794 on RAD-ChestCT.
Beyond classification, \ours-CT enables zero-shot free-text segmentation on 3D CT volumes (Table~\ref{tab:ct_freetext_segmentation}). On ReXGroundingCT, \ours-CT attains 0.128 Dice, on par with SAT~\cite{sat} (0.131), even though SAT and VoxTell~\cite{voxtell} are fine-tuned on ReXGroundingCT while \ours-CT uses no mask supervision. To our knowledge, this is the first zero-shot free-text segmentation on 3D CT to match supervised methods.

\textbf{Chest CT downstream evaluation.}
We compare \ours-CT with prior chest CT VLMs~\cite{ctclip, fvlm, visd-boost, colipri} and V-JEPA 2.1~\cite{vjepa21} (Table~\ref{tab:ct_downstream}); fVLM and ViSD-Boost are excluded from segmentation evaluation due to their dependence on organ-level segmentation.
\ours-CT achieves the best performance on the fine-grained, lesion-level benchmarks: RSNA-PE (pulmonary embolism), LIDC-IDRI (pulmonary nodules), MSD-Lung (primary lung cancers), and NSCLC-Radiomics (lung tumors). On CT-RATE, it ties COLIPRI (0.876).
The vision-language alignment training also raises performance substantially over its V-JEPA 2.1 initialization (e.g., 0.788 to 0.901 on LIDC-IDRI, 0.532 to 0.613 on MSD-Lung), showing that \ours\ produces strong, transferable representations particularly suited to fine-grained CT tasks.
For report generation on CT-RATE, \ours-CT achieves the best clinical efficacy (Macro F1 and CRG), while METEOR and ROUGE-L trail the best baseline results; this suggests that our representations capture clinical content well, with room for improvement on surface-level fluency.

\subsection{Ablation Study}
\label{sec:ablation}

We ablate the components of \ours\ in Table~\ref{tab:ablation}, reporting task-wise averages across each evaluation suite.
Comparing rows (1) and (2), unfreezing the SSL encoder yields large zero-shot gains, especially on dense localization, where CXR segmentation rises from 0.224 to 0.326, confirming the importance of VL adaptation.
Replacing the VL-CABS~\cite{radzero} architecture with our SGA in row (3) preserves global classification while substantially improving segmentation (CXR from 0.326 to 0.369; CT from 0.084 to 0.117), consistent with SGA's sparsity over a separate gate embedding space.
Adding the gate confidence shift or DFR alone yields mixed effects, but combining both in row (6) is the strongest, with the largest gains on localization-heavy tasks (CXR grounding 0.810; CT segmentation 0.128); the two are complementary, with DFR preserving the fine-grained features needed for localization and the gate confidence shift refining the alignment signal.
Section~\ref{sec:sparse_precise} analyzes this complementarity
at the patch level via sparsity and precision.
Appendix~\ref{apx:additional_ablations} reports further ablations on the DFR design, CXR input resolution and encoder size, and pretrained initialization with teacher regularization.

\input{tables/ablation}

\begin{figure}[t]
  \centering
  \includegraphics[width=\textwidth]{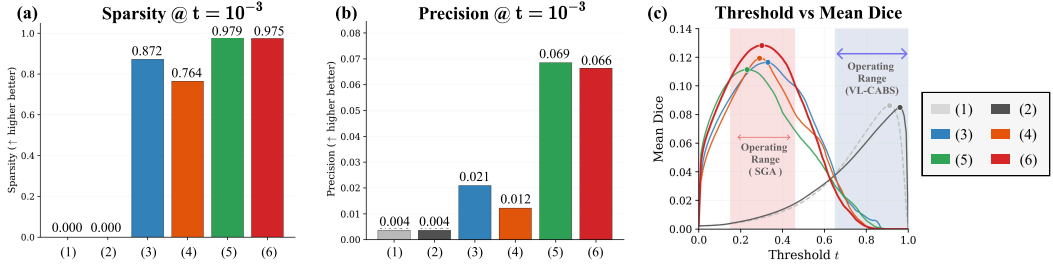}
\caption{Sparse and precise activation for ablation variants (1)--(6). \textbf{(a)} Patch-wise sparsity $\Pr(m_p < \theta)$ and \textbf{(b)} precision $TP/(TP+FP)$ at $\theta\!=\!10^{-3}$; \textbf{(c)} voxel-wise Dice across thresholds.}
  \label{fig:sparse_precise}
\end{figure}

\subsubsection{Analysis for Sparse and Precise Activation}
\label{sec:sparse_precise}

To understand how \ours\ translates patch-level sparsity into
voxel-level localization, we report three metrics on ReXGroundingCT
(Figure~\ref{fig:sparse_precise}): patch-wise sparsity
$\Pr(m_p < \theta)$ and precision $TP/(TP+FP)$ at a threshold of $\theta\!=\!10^{-3}$,
and voxel-wise Dice across thresholds after trilinear upsampling,
where $m_p$ is the patch-wise localization score (Section~\ref{sec:inference}) or, for VL-CABS~\cite{radzero}, the patch-level similarity score.

The VL-CABS baselines (1, 2) activate every patch (sparsity $0.000$) with precision equal to the foreground ratio ($0.004$). 
Replacing VL-CABS with SGA (3) suppresses $87.2\%$ of patches, and adding DFR (5) further raises this to $97.9\%$, approaching the background ratio of $99.6\%$. 
This sparsity contributes to precision: variant (6) reaches substantially higher precision than VL-CABS ($0.066$ vs.\ $0.004$). The voxel-wise Dice profiles also differ. VL-CABS baselines peak only near $\theta\!=\!1$, where high thresholds finally separate 
foreground from background. The SGA variants, already sparse at low thresholds ($\theta\!=\!10^{-3}$), peak at intermediate thresholds ($\theta$ between $0.2$ and $0.4$).

Each component alone, in (4) and (5), trades sparsity against Dice. DFR (5) attains the highest sparsity but the lowest Dice ($0.112$) among SGA variants, as the regularization toward SSL features may also suppress patches that still carry finding-relevant signal. 
The gate confidence shift, in contrast, incorporates the pooled gate confidence as an additional logit signal; it slightly relaxes sparsity but improves Dice ($0.119$ in (4)). The full configuration (6) combines both and resolves this trade-off. DFR enforces near-maximal sparsity ($0.975$) while the gate confidence shift recovers the discriminative signal, yielding a precision of $0.066$ and the best voxel-wise Dice ($0.128$ vs.\ $0.086$ for the frozen baseline (1)).

\subsection{Visualization}
\label{sec:visualization}

\begin{figure}[t]
    \begin{center}
    \includegraphics[width=\linewidth]{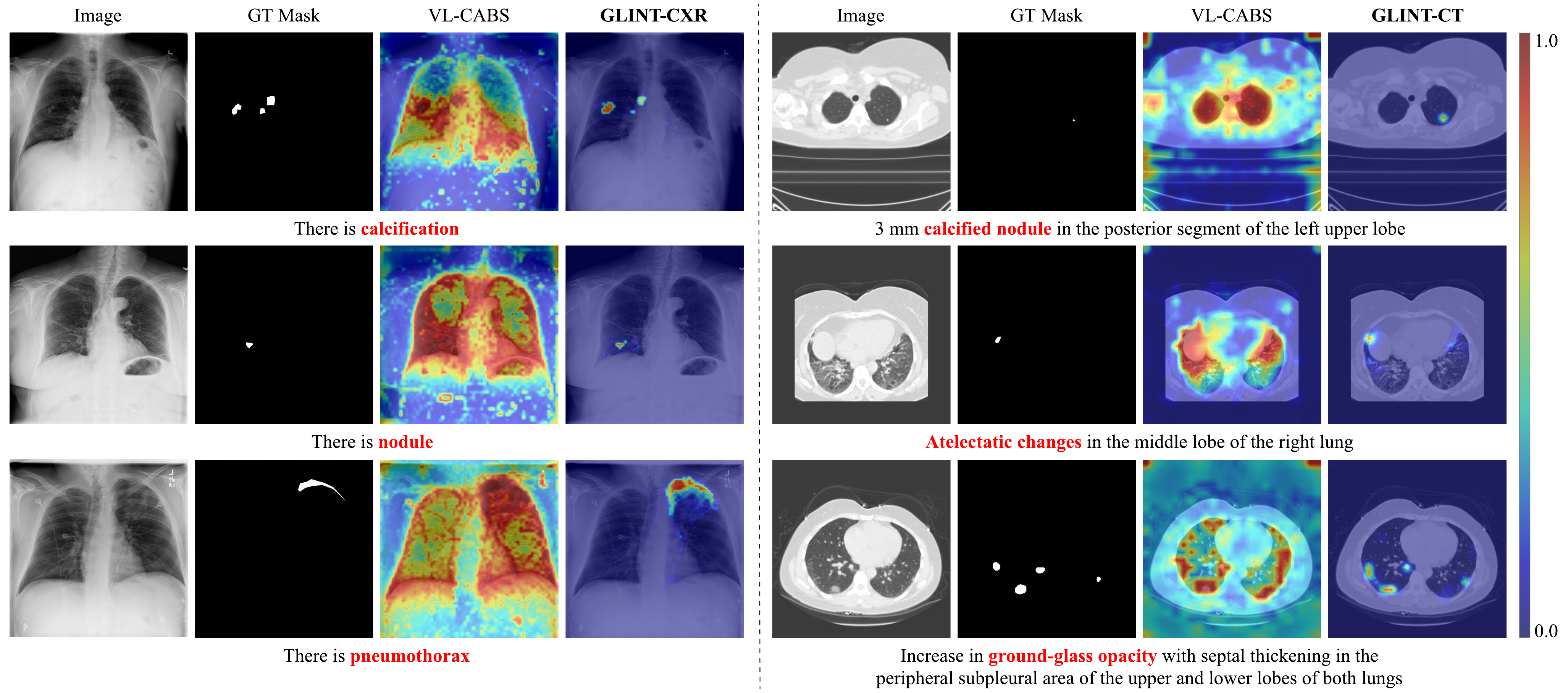}
    \caption{Visualization of the \emph{Gated Similarity Map} (GSM) on chest X-ray (left) and chest CT (right). For each sentence, we compare \ours\ with VL-CABS~\cite{radzero}. \ours\ produces sparse activations concentrated on the finding, whereas VL-CABS spreads alignment across irrelevant regions.}
    \label{fig:visualization}
    \end{center}
\end{figure}

We qualitatively visualize the \emph{Gated Similarity Map} (GSM) of \ours\ in Figure~\ref{fig:visualization} and compare against VL-CABS~\cite{radzero} (ablation~(2) in Table~\ref{tab:ablation}) on ChestX-Det~\cite{chestXdet} and ReXGroundingCT~\cite{rexgroundingct}.
Across both modalities, \ours\ accurately captures the location and size of each finding and closely aligns with the corresponding text phrase, even for fine-grained lesions or free-text descriptions.
These patterns demonstrate that the design of \ours\ for sparse alignment effectively concentrates vision-language patch-level similarity on finding-relevant regions, consistent with the analysis in Section~\ref{sec:sparse_precise}.
Additional visualization results of findings and PCA analyses are provided in Appendix~\ref{apx:visualization}.

%% file: tables/ct_zeroshot.tex
\begin{table*}[!t]
\centering

\begin{minipage}[t]{0.48\textwidth}
\centering
\small
  \caption{
    Zero-shot performance on chest CT.
  }
  \centering
  \small
  \setlength{\tabcolsep}{6pt}
  \renewcommand{\arraystretch}{1.0}

\resizebox{0.8\columnwidth}{!}{%
    \begin{NiceTabular}{l *{2}{wc{1.5cm}}}[color-inside]
      \toprule
      \Block[c]{3-1}{Method}
        & \multicolumn{2}{c}{\textbf{Classification}} \\
      \specialrule{0pt}{1pt}{1pt}

        & CT-RATE & RAD-ChestCT \\
      \specialrule{0pt}{1pt}{1pt}

        & \multicolumn{2}{c}{Macro AUROC} \\
      \specialrule{0pt}{1pt}{0pt}
      \toprule

      CT-CLIP~\cite{ctclip}                     & 0.731 & 0.629 \\
      Merlin$^\dagger$~\cite{merlin}            & 0.728 & 0.644 \\
      fVLM$^\dagger$~\cite{fvlm}                & 0.778 & 0.680 \\
      ViSD-Boost$^\dagger$~\cite{visd-boost}    & 0.790 & 0.694 \\
      RadZero3D~\cite{radzero3d}                & 0.762 & 0.690 \\
      COLIPRI~\cite{colipri}                    & 0.785 & 0.730 \\

      \specialrule{0pt}{0pt}{1pt}
      \rowcolor{gray!20}
      \ours-CT                                  & \textbf{0.850} & \textbf{0.789} \\
      \specialrule{0.7pt}{0pt}{-1pt}

      \CodeAfter
      \begin{tikzpicture}
        \begin{scope}[shorten <=2pt, shorten >=2pt]
          \draw (1-|2) -- (4-|2);
          \draw (4-|2) -- (11-|2);
        \end{scope}
        \begin{scope}[shorten <=2.5pt, shorten >=2.5pt]
          \draw (2-|2) -- (2-|4);
          \draw (3-|2) -- (3-|4);
        \end{scope}
      \end{tikzpicture}
    \end{NiceTabular}%
  }
  \label{tab:ct_zeroshot}
\end{minipage}%
\hfill
\begin{minipage}[t]{0.48\textwidth}
\centering
\small
  \caption{
    Zero-shot free-text segmentation on ReXGroundingCT.
    For reference, supervised methods are reported.
    $^\ddagger$ are from \cite{voxtell}.
  }
  \centering
  \small
  \setlength{\tabcolsep}{4pt}
  \renewcommand{\arraystretch}{1.0}

  \resizebox{\columnwidth}{!}{%
    \begin{NiceTabular}{p{2.2cm} c c}[color-inside]
      \toprule
      \Block[c]{3-1}{Method} & \Block[c]{3-1}{Setting} & \textbf{Free-text Segmentation} \\
      \specialrule{0pt}{1pt}{1pt}
                              &                         & ReXGroundingCT \\
      \specialrule{0pt}{1pt}{1pt}
                              &                         & Dice \\
      \specialrule{0pt}{1pt}{0pt}
      \toprule

      SAT$^\ddagger$~\cite{sat}                  & Supervised   & 0.131 \\
      VoxTell$^\ddagger$~\cite{voxtell}          & Supervised   & 0.282 \\

      \specialrule{0pt}{0pt}{1pt}
      \rowcolor{gray!20}
      \ours-CT                        & Zero-shot    & 0.128 \\
      \specialrule{0.7pt}{0pt}{-1pt}

      \CodeAfter
      \begin{tikzpicture}
        \begin{scope}[shorten <=2pt, shorten >=2pt]
          \draw (1-|3) -- (4-|3);   
          \draw (4-|3) -- (7-|3);   
        \end{scope}
        \begin{scope}[shorten <=3pt, shorten >=3pt]
          \draw (2-|3) -- (2-|4);   
          \draw (3-|3) -- (3-|4);   
        \end{scope}
      \end{tikzpicture}
    \end{NiceTabular}%
  }
  \label{tab:ct_freetext_segmentation}
\end{minipage}

\end{table*}
  

%% file: tables/ct_downstream.tex
\begin{table*}[!t]
\centering
\small
\caption{
  Downstream task performance on chest CT,
  averaged over three random seeds.
  Bold denotes the highest mean performance.
}
  \centering
  \small
  \setlength{\tabcolsep}{2pt}
  \renewcommand{\arraystretch}{1.15}
  \resizebox{\textwidth}{!}{%
    \begin{NiceTabular}{p{2.2cm} *{3}{wc{2.0cm}} *{4}{wc{1.8cm}} *{2}{wc{1.7cm}}}[color-inside]
      \toprule
      \Block[c]{3-1}{Method}
      & \multicolumn{3}{c}{\textbf{Classification}}
      & \multicolumn{4}{c}{\textbf{Report Generation}}
      & \multicolumn{2}{c}{\textbf{Segmentation}} \\
      \specialrule{0pt}{2pt}{3pt}
      & CT-RATE & RSNA-PE & LIDC-IDRI
      & \multicolumn{4}{c}{CT-RATE}
      & MSD-Lung & NSCLC-Rad \\
      \specialrule{0pt}{2pt}{3pt}
      & Macro AUROC
      & \multicolumn{2}{c}{AUROC}
      & METEOR & ROUGE-L & RadBERT F1 & CRG
      & \multicolumn{2}{c}{Dice} \\
      \specialrule{0pt}{2pt}{0pt}
      \toprule
      V-JEPA 2.1~\cite{vjepa21}
      & 0.857 & 0.627 & 0.788 & 0.223 & 0.271 & 0.338 & 0.397 & 0.532 & 0.561 \\
      CT-CLIP~\cite{ctclip}
      & 0.766 & 0.595 & 0.786 & 0.191 & 0.274 & 0.124 & 0.352 & 0.022 & 0.360 \\
      fVLM~\cite{fvlm}
      & 0.807 & 0.649 & 0.730 & 0.224 & 0.290 & 0.327 & 0.394 & -- & -- \\
      ViSD-Boost~\cite{visd-boost}
      & 0.801 & 0.601 & 0.827 & 0.219 & \textbf{0.294} & 0.293 & 0.384 & -- & -- \\
      COLIPRI~\cite{colipri}
      & \textbf{0.876} & 0.673 & 0.874 & \textbf{0.228} & 0.289 & 0.391 & 0.425 & 0.568 & 0.581 \\
      \specialrule{0pt}{0pt}{1pt}
      \rowcolor{gray!20}
      \ours-CT
      & \textbf{0.876} & \textbf{0.677} & \textbf{0.901} & 0.221 & 0.226 & \textbf{0.399} & \textbf{0.432} & \textbf{0.613} & \textbf{0.587} \\
      \specialrule{0.7pt}{0pt}{-1pt}
      \CodeAfter
      \begin{tikzpicture}
          \begin{scope}[shorten <=2pt, shorten >=2pt]
            \draw (1-|2) -- (4-|2);
            \draw (1-|5) -- (4-|5);
            \draw (1-|9) -- (4-|9);
        \end{scope}
      \begin{scope}[shorten <=2pt, shorten >=2pt]
            \draw (4-|2) -- (10-|2);
            \draw (4-|5) -- (10-|5);
            \draw (4-|9) -- (10-|9);
        \end{scope}
        \begin{scope}[shorten <=3pt, shorten >=3pt]
            \draw (2-|2) -- (2-|5);   
            \draw (2-|5) -- (2-|9);   
            \draw (2-|9) -- (2-|11);  
            \draw (3-|2) -- (3-|3);
            \draw (3-|3) -- (3-|5);
            \draw (3-|5) -- (3-|9);
            \draw (3-|9) -- (3-|11);
        \end{scope}
      \end{tikzpicture}
    \end{NiceTabular}%
    }
  \label{tab:ct_downstream}
\end{table*}

%% file: tables/ablation.tex
\begin{table*}[!t]
  \caption{
    Ablation results with task-wise averages on chest X-ray and chest CT. Results use the same evaluation metrics as in the main performance tables.
  }
  \centering
  \small
  \setlength{\tabcolsep}{5pt}
  \renewcommand{\arraystretch}{1.15}

  \resizebox{\textwidth}{!}{
    \begin{NiceTabular}{l wc{1.55cm} wc{1.7cm} *{2}{wc{1.55cm}} *{5}{wc{1.55cm}}}[color-inside]
      \toprule

      \Block[c]{2-1}{}
        & \Block[c]{2-1}{\textbf{Trainable Encoder}}
        & \Block[c]{2-1}{\textbf{VL\\Architecture}}
        & \Block[c]{2-1}{\textbf{Gate Confidence Shift}}
        & \Block[c]{2-1}{\textbf{DFR}}
        & \multicolumn{3}{c}{\textbf{CXR Zero-shot}}
        & \multicolumn{2}{c}{\textbf{Chest CT Zero-shot}} \\
      \specialrule{0pt}{2pt}{3pt}

        & & & &
        & Classification & Grounding & Segmentation
        & Classification & Segmentation \\
      \specialrule{0pt}{2pt}{0pt}
      \toprule

      (1)
          & --        & VL-CABS~\cite{radzero}       & --        & --
          & 0.814 & 0.680 & 0.224 & 0.785 & 0.086 \\

      (2)
          & \ding{51} & VL-CABS~\cite{radzero}       & --        & --
          & 0.866 & 0.792 & 0.326 & 0.818 & 0.084 \\

      (3)
          & \ding{51} & SGA & --        & --
          & 0.866 & 0.781 & 0.369 & \textbf{0.819} & 0.117 \\

      (4)
          & \ding{51} & SGA & \ding{51} & --
          & 0.870 & 0.772 & 0.369 & 0.812 & 0.119 \\

      (5)
          & \ding{51} & SGA & --        & \ding{51}
          & 0.871 & 0.787 & 0.364 & 0.810 & 0.112 \\

      \specialrule{0pt}{0pt}{1pt}
      \rowcolor{gray!20}
      (6) \ours
          & \ding{51} & SGA & \ding{51} & \ding{51}
          & \textbf{0.877} & \textbf{0.810} & \textbf{0.375} & \textbf{0.819} & \textbf{0.128} \\
      \specialrule{0.7pt}{0pt}{-1pt}

      \CodeAfter
      \begin{tikzpicture}
        \begin{scope}[shorten <=2pt, shorten >=2pt]
          \draw (1-|6) -- (3-|6);
          \draw (1-|9) -- (3-|9);
          \draw (3-|6) -- (9-|6);
          \draw (3-|9) -- (9-|9);
        \end{scope}

        \begin{scope}[shorten <=3pt, shorten >=3pt]
          \draw (2-|6) -- (2-|9);    
          \draw (2-|9) -- (2-|11);   
        \end{scope}
      \end{tikzpicture}

    \end{NiceTabular}
  }
  \label{tab:ablation}
\end{table*}

%% file: sec/6_conclusion.tex
\section{Conclusion}
\label{sec:conclusion}

We present \ours, a vision-language framework that addresses the sparse image-report correspondence in radiology along two complementary axes.
SGA achieves sparsity by activating, for each textual query, only the patches that are actually relevant, via a sigmoid gate over a separate gate embedding space.
DFR preserves the fine-grained features of self-supervised foundations during VL training by anchoring intermediate features to a frozen SSL teacher.
The same recipe generalizes across both 2D CXR and 3D CT, yielding \ours-CXR and \ours-CT.
On chest X-ray and chest CT, \ours\ achieves strong zero-shot classification, grounding, and segmentation from free-text queries, including, to our knowledge, the first zero-shot free-text segmentation on 3D CT, and its representations outperform both the underlying SSL encoders and prior medical vision-language models on downstream classification, report generation, and segmentation.

\textbf{Limitations and future work.}
Our evaluation focuses on chest X-ray and chest CT; extending the same recipe to other modalities and anatomical regions (e.g., abdominal CT, brain MRI) is left for future work.
Beyond benchmark evaluation, direct clinician assessment of \ours's clinical utility in practice remains an important next step.

\begin{ack}
This work was supported by the Technology Innovation Program
(RS-2025-02221011, Development of Medical-Specialized Multimodal
Hyperscale Generative AI Technology for Global Integration)
funded by the Ministry of Trade, Industry \& Energy
(MOTIE, South Korea), and by the Advanced GPU Utilization
Support Program funded by the Korean government
(Ministry of Science and ICT) (Project No. 02-26-91-0330).
\end{ack}

%% file: sec/X_appendix.tex
\newpage

\section{Additional Visualization Results}
\label{apx:visualization}

\begin{figure}[b!]
    \begin{center}
    \includegraphics[width=\linewidth]{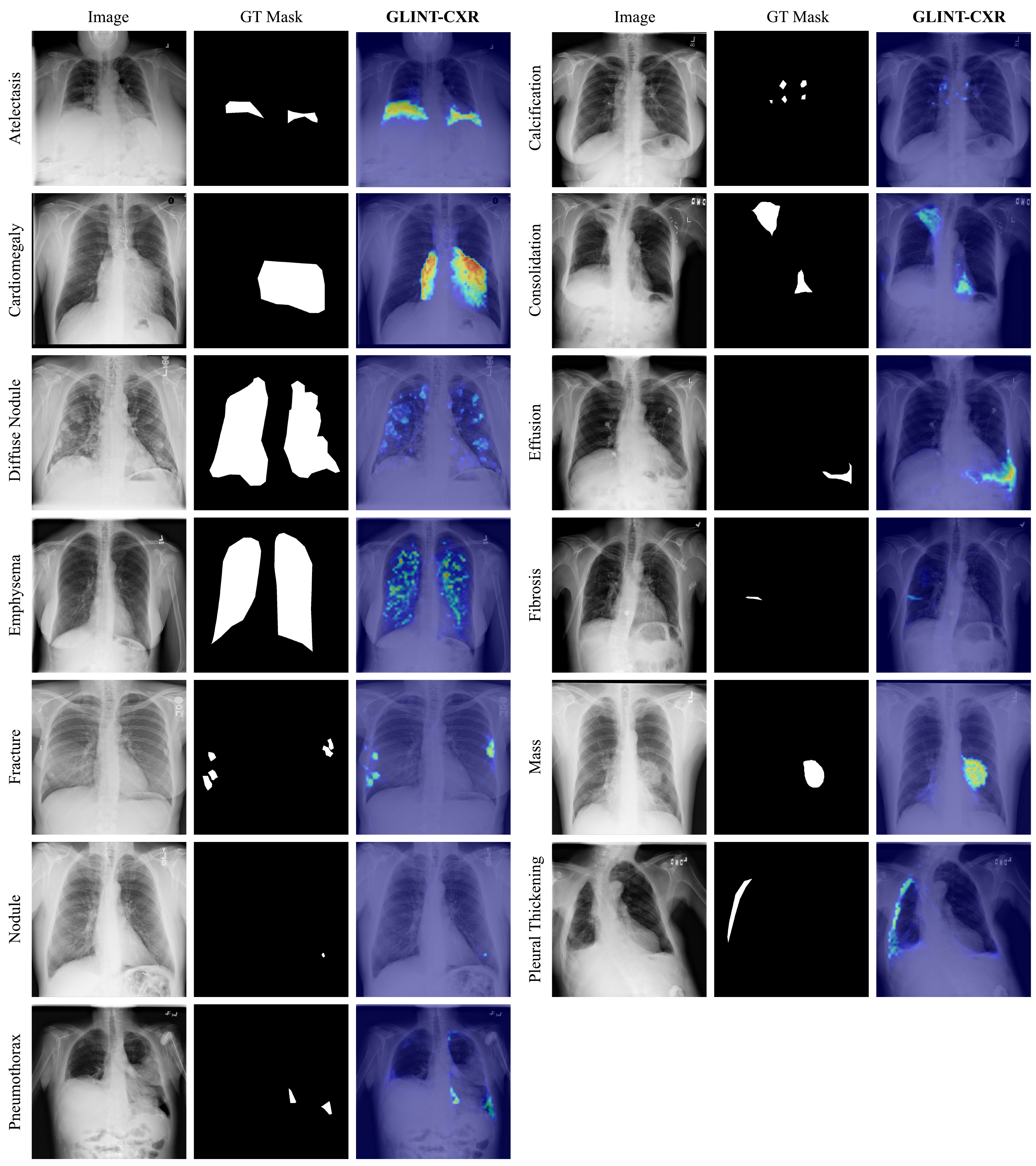}
    \caption{Visualization of Gated Similarity Maps (GSM) on chest X-ray for all 13 findings in ChestX-Det~\cite{chestXdet}.}
    \label{fig:appendix_cxr_visualization}
    \end{center}
\end{figure}

\begin{figure}[t]
    \begin{center}
    \includegraphics[width=\linewidth]{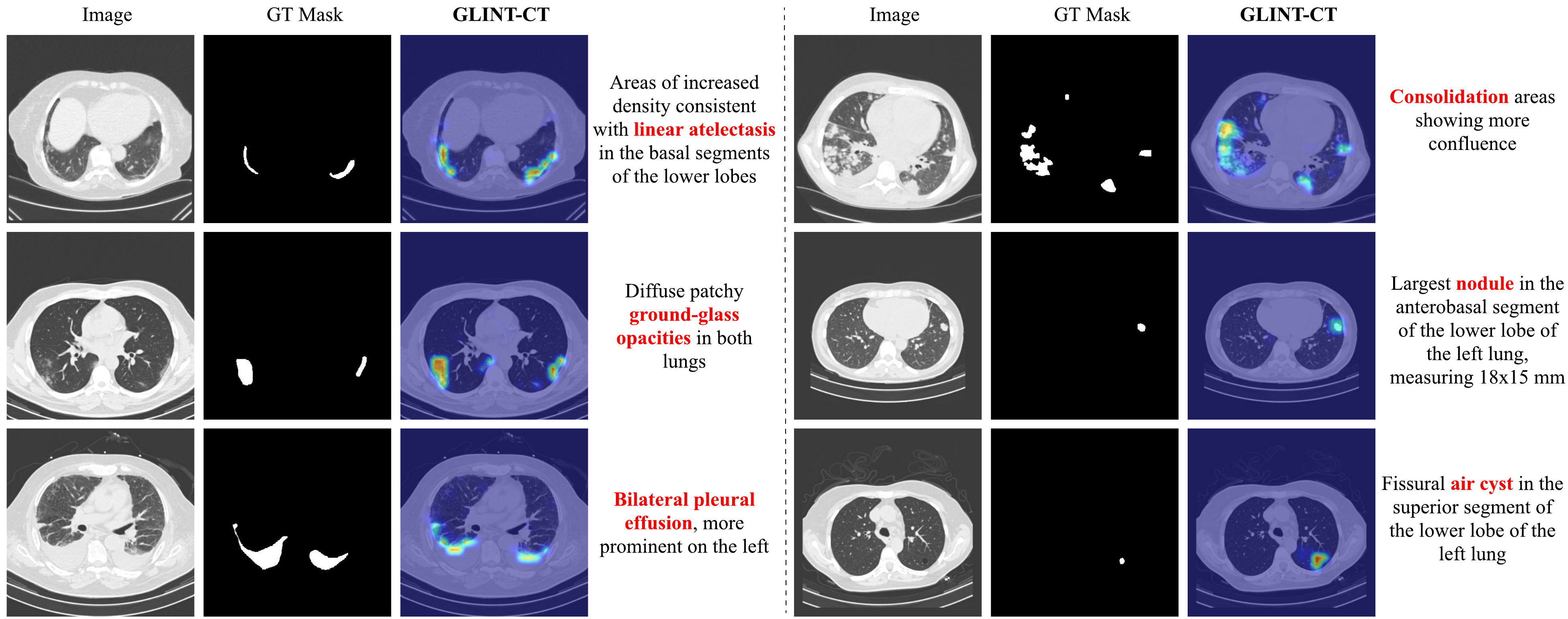}
    \caption{Visualization of Gated Similarity Maps (GSM) on chest CT for several findings from ReXGroundingCT~\cite{rexgroundingct}.}
    \label{fig:appendix_ct_visualization}
    \end{center}
\end{figure}

\paragraph{Gated Similarity Maps (GSM).}
We provide GSM visualizations for both modalities.
Figure~\ref{fig:appendix_cxr_visualization} covers all 13 findings in ChestX-Det~\cite{chestXdet} on chest X-ray, and Figure~\ref{fig:appendix_ct_visualization} covers several findings from ReXGroundingCT~\cite{rexgroundingct} on chest CT.
For each case, we present the input image (axial slice for CT), the ground-truth mask, and the GSM produced by \ours\ for the corresponding text phrase.

\begin{figure}[t]
    \begin{center}
    \includegraphics[width=\linewidth]{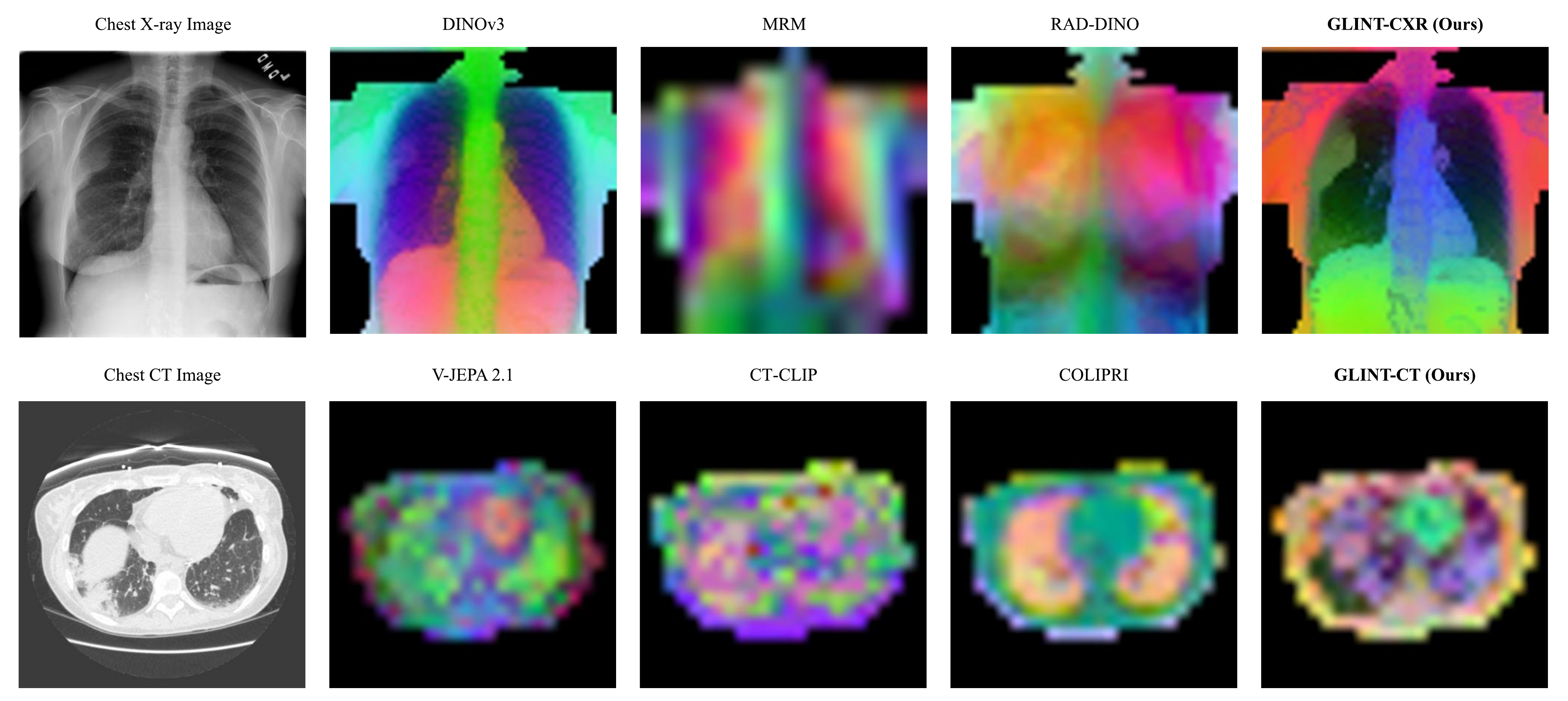}
    \caption{PCA maps of dense patch-level features on chest X-ray (top) and chest CT (bottom).}
    \label{fig:appendix_pca_visualization}
    \end{center}
\end{figure}

\paragraph{Principal component analysis (PCA).}
To examine the patch-level feature space, Figure~\ref{fig:appendix_pca_visualization} visualizes PCA maps of dense patch features on chest X-ray (top) and chest CT (bottom).
The CXR example is a lung mass case from ChestX-Det~\cite{chestXdet}, and the CT example is a lung cancer case from MSD-Lung~\cite{antonelli2022medical}.
For each modality, we compare \ours\ with its underlying SSL backbone (DINOv3~\cite{dinov3} for CXR, V-JEPA~2.1~\cite{vjepa21} for CT) and representative baselines.
Each image is processed at each model's native input resolution and patch size, and the resulting feature maps are visualized with bicubic interpolation.
\ours\ produces sharper and more coherent features and more clearly delineates the lesion region, indicating that it captures fine-grained, lesion-aware representations.

\section{Implementation Details}
\label{apx:implementation_detail}

\paragraph{Chest CT preprocessing.}
Chest CT volumes are preprocessed following CT-CLIP~\cite{ctclip}, implemented with MONAI~\cite{monai} so that the pipeline can be applied uniformly across CT-RATE, RAD-ChestCT, RSNA-PE, and LIDC-IDRI.
Volumes are loaded and reoriented to the IPL orientation, then resampled to a target voxel spacing of $(1.5, 0.75, 0.75)\,\mathrm{mm}$ in $(D, H, W)$ via trilinear interpolation.
Hounsfield units are clipped to $[-1000, 1000]$ and linearly mapped to $[-1, 1]$.
Volumes are center-cropped or zero-padded to a region-of-interest size of $(240, 480, 480)$, and then trilinearly resized to a final size of $(240, 384, 384)$ to match the input resolution of V-JEPA 2.1~\cite{vjepa21}.
To construct pseudo-RGB inputs, consecutive slices are stacked to convert $(D, H, W)$ volumes into $(D/3, 3, H, W)$.
Values are then rescaled from $[-1, 1]$ to $[0, 1]$, followed by the standard V-JEPA 2.1 input z-normalization.

\paragraph{Gate head.}
Both $\phi_t$ and $\phi_v$ are 2-layer MLPs with GELU activations~\cite{gelu}.
The hidden dimension is set to 768 for CXR and chest CT.

\paragraph{Dense feature regularization.}
The projector $h$ is a 3-layer MLP with SiLU activations~\cite{silu}.
The hidden dimension is set to 4,096 for CXR and 2,048 for chest CT.
Since the teacher encoder is frozen, we precompute its patch-level features once and cache them on disk, loading the cached features during training to reduce GPU memory.

\input{tables/appendix_training_details}

\paragraph{Training details.}
We use the cosine warm-up scheduler with 50 warm-up steps and a weight decay of 0.05, and apply an early stopping strategy for all training.
All other
task-specific settings are reported in Table~\ref{tab:appendix_training_details}.
Pretraining is conducted using automatic mixed precision with bfloat16, while all downstream tasks are trained in full single-precision.

\paragraph{Downstream task details.}
For \textbf{classification}, we adopt attentive probing, where an attentive pooling layer followed by a linear classifier is trained on top of the frozen features.
For \textbf{report generation}, an R2Gen~\cite{r2gen} decoder is attached to the frozen encoder.
For \textbf{CXR segmentation}, a simple UPerNet~\cite{upernet}-style decoder is adopted.
To use the same decoder architecture across different vision encoders, patch-level features are adaptively pooled to a spatial resolution of $16^2$ before decoding.
For \textbf{CT segmentation}, input volumes are preprocessed to match each encoder's input format, and a UNETR~\cite{unetr2022}-style 3D decoder without skip connections is used to produce predictions at the input volume size. Inference is performed using MONAI's sliding-window inference~\cite{monai} with Gaussian importance weighting and $0.5$ overlap, following nnU-Net conventions~\cite{isensee2021nnunet}.

\paragraph{Computational cost.}
Table~\ref{tab:compute_cost} summarizes the computational cost of training \ours-CXR and \ours-CT.
\input{tables/computational_cost}

\section{Report Decomposition Prompt Design}
\label{sec:appendix_prompts}

For each report, we extract the Findings and Impression sections and use them as input to the decomposition prompt, which is applied uniformly to both CXR and chest CT reports.
The prompt converts the free-text input into clinical sentences.
Temporal comparisons are removed, and all statements are rewritten as present-tense declarative sentences following consistent templates (e.g., ``There is ...'', ``There is no ...'', ``There may be ...'').
Sentences containing multiple findings connected by commas or conjunctions (``and'', ``or'', ``with'') are split into one sentence per finding.
Both positive and negative statements are explicitly included; normal observations (e.g., ``the heart size is normal'') are kept as declarative sentences rather than discarded.
Uncertainty expressions are preserved, and for statements containing diagnostic phrasing (e.g., ``to suggest pneumonia''), both complete and observation-only variants are extracted.

\input{tables/appendix_test_dataset}

\section{Evaluation Dataset Details}
\label{apx:test_dataset}

\subsection{Chest X-ray Zero-shot}
To enable a fair comparison across models, we follow the evaluation protocol of RadZero~\cite{radzero}\footnote{ \url{https://huggingface.co/Deepnoid/RadZero/blob/main/data/README.md}}, using the same images and data splits.
For zero-shot segmentation, we additionally evaluate on positive cases from \textbf{ChestX-Det}~\cite{chestXdet} (13 classes).
Since MS-CXR~\cite{mscxr} is built on MIMIC-CXR, we exclude its test images from our MIMIC-CXR training data to prevent leakage.

\subsection{Chest X-ray Downstream}
\textbf{ChestXray14}~\cite{chestxray14} provides official training and test sets.
We split the former at a 9:1 ratio to obtain training and validation sets, and use the latter for evaluation.
\textbf{VinDr-CXR}~\cite{vindrcxr} provides official training and test sets.
The former is randomly split into training and validation sets with a ratio of 9:1, while the latter is used for evaluation.
VinDr-CXR provides 28 classes in total. After excluding ``Edema'', which is not present in the test set, we perform multi-label classification over the remaining 27 classes.
\textbf{RSNA Pneumonia (RSNA-PN)}~\cite{rsna} provides labels only for its official training set.
We use the labeled training set and partition it at the patient level into training, validation, and test sets with a ratio of 8:1:1.
For \textbf{MIMIC-CXR}~\cite{mimiccxr}, we follow the official training, validation, and test sets, and include
only reports in which the Findings section is present.
For computational reasons, we adopt a random 10\% subset of the training set, while the validation and test sets remain unchanged.
For \textbf{SIIM-ACR Pneumothorax}~\cite{siim}, we use a dataset version that provides PNG-format images with image-level segmentation masks, enabling clear separation between splits\footnote{ \url{https://www.kaggle.com/datasets/jesperdramsch/siim-acr-pneumothorax-segmentation-data}}.
Following the data split defined in BenchX~\cite{benchx}, we adopt training, validation, and test sets with a ratio
of 7:1.5:1.5. 
\textbf{ChestX-Det}~\cite{chestXdet} is released with official training and test sets.
The former is split into training and validation sets with a ratio of 9:1, while the latter is used for evaluation.

\subsection{Chest CT Zero-shot}
The official validation split of \textbf{CT-RATE}~\cite{ctclip} serves as the internal evaluation and \textbf{RAD-ChestCT}~\cite{radchestct} is used for external evaluation.
While RAD-ChestCT originally provides 84 abnormality labels, we use 16 mapped abnormalities following CT-CLIP~\cite{ctclip}.
As the label mapping procedure is not specified in CT-CLIP~\cite{ctclip}, we employ the mapping strategy from RadZero3D~\cite{radzero3d}.
For free-text segmentation, we use the 50-case validation split of \textbf{ReXGroundingCT}~\cite{rexgroundingct}.
Since ReXGroundingCT is built on CT-RATE, we exclude these validation volumes from our CT-RATE training data to prevent leakage.

\subsection{Chest CT Downstream}
Since fVLM and ViSD-Boost require masks to crop regions of interest from CT volumes, we apply the same preprocessing pipeline to \textbf{RSNA-PE}~\cite{rsna_pe} and \textbf{LIDC-IDRI}~\cite{lidc}. Following their pipeline, masks are generated using TotalSegmentator~\cite{totalsegmentator}, and samples with failed segmentation are excluded.
For a fair comparison, although \ours-CT and CT-CLIP do not rely on mask-based preprocessing, evaluation is restricted to the intersection of samples available across all methods.

Dataset split strategies vary across datasets.
For \textbf{CT-RATE}, the official validation split
(3,002 volumes from 1,304 patients) serves as our test set.
To establish a validation set during training, we extract a subset
of 1,304 patients from the 20,000 patients in the official training
split of CT-RATE, matching the number of patients in its official
validation split.
The remaining 18,696 patients form the training set, and all scans
and reconstructions from each patient are kept in the same split.

After excluding 803 non-chest or unusable volumes, the training,
validation, and test sets contain 43,390, 2,993, and 3,002 volumes,
respectively.
For vision-language alignment, we further exclude volumes whose
report decomposition yields no sentences and volumes overlapping
ReXGroundingCT from the training and validation sets, leaving
43,337 and 2,990 volumes, respectively.
All 3,002 test volumes are retained for every task.
In the case of RSNA-PE, official splits are provided, but the test set lacks labels.
The official training split is therefore divided at the study level into train, validation, and test sets with a ratio of 7:1.5:1.5, following~\cite{ke2024video}.
As LIDC-IDRI does not provide official splits, we use patient-disjoint
training, validation, and test sets containing 705, 148, and 157 scans,
respectively. For segmentation, \textbf{MSD-Lung}~\cite{antonelli2022medical} and \textbf{NSCLC-Radiomics}~\cite{aerts2014decoding} serve as evaluation benchmarks.
Since neither dataset provides official splits, we partition both at the patient level into train, validation, and test sets with a 7:1:2 ratio.
For NSCLC-Radiomics, the model is trained on two foreground classes (primary and secondary GTV). However, as secondary GTV annotations are available for only a subset of patients, we evaluate Dice on the primary GTV alone.

\section{Additional Ablation Studies}
\label{apx:additional_ablations}

We provide three additional ablations beyond the main results:
(i)~the design choices of DFR,
(ii)~the input resolution and vision encoder size for chest X-ray,
and (iii)~pretrained initialization and teacher regularization.

\subsection{Dense Feature Regularization}
\label{apx:dfr_ablation}
We analyze DFR along three axes: (i)~whether explicit distillation is needed, by replacing DFR with a partially frozen encoder baseline (all but the last two student layers frozen, no DFR); (ii)~teacher capacity, by replacing the large SSL teacher (ViT-7B for CXR, ViT-G for CT) with a student-sized ViT-L; and (iii)~feature extraction depth, sweeping $50\%$, $75\%$, $100\%$ against our ${\sim}90\%$ setting ($(l_s,l_t)=(22,36)$ for CXR, $(22,44)$ for CT). Results are in Table~\ref{tab:appendix_dfr_ablation}.

The partially frozen baseline drops on every metric, showing that freezing alone is insufficient. DFR fills this gap: a student-sized ViT-L teacher already recovers much of the drop, and the ViT-7B/G teacher yields further gains, suggesting that the SSL anchor drives most of the improvement, with a larger teacher contributing additional gains.
Across depth, \ours\ ranks first on $4$ of $5$ metrics. The $100\%$ setting yields the lowest CXR segmentation and trails \ours-CT on both CT metrics, suggesting too little capacity for VL adaptation; $50$--$75\%$ underperforms broadly, consistent with weak regularization in the upper layers. ${\sim}90\%$ balances these regimes across CXR and CT.

\input{tables/appendix_dfr_ablation}
\input{tables/appendix_cxr_resolution_ablation}

\subsection{Input Resolution and Vision Encoder Size (CXR)}
\label{apx:cxr_resolution_ablation}

Existing CXR baselines (e.g., RadZero~\cite{radzero}) typically use ${\sim}518^2$ inputs with a ViT-B backbone, whereas \ours-CXR uses $1024^2$ with ViT-L. To verify that our gains are not driven by larger inputs or backbone alone, we conduct a controlled ablation over $\{512^2, 1024^2\} \times \{\text{ViT-B}, \text{ViT-L}\}$. Results are in Table~\ref{tab:appendix_cxr_resolution_ablation}.

Even at the matched $512^2{+}\text{ViT-B}$ configuration, our framework exceeds prior baselines (Table~\ref{tab:chest_xray_zeroshot}) on 9 of 10 metrics, indicating that the gains stem from the framework, not solely from resolution or backbone. \ours-CXR ($1024^2{+}\text{ViT-L}$) improves further, ranking first on 8 of 10 metrics.

\subsection{Initialization and Teacher Ablations}
\label{apx:initialization}

\paragraph{Pretrained initialization and DFR.}
We examine the effects of pretrained student initialization and DFR
in Table~\ref{tab:appendix_initialization_dfr}, reporting task-wise
averages as in Table~\ref{tab:ablation}.
With pretrained initialization, the model without DFR still
outperforms the best prior results in all four task groups with
available baselines.
Removing pretrained initialization substantially degrades performance,
underscoring its importance.
DFR improves all five task groups even from scratch, indicating
that initialization and regularization provide complementary benefits.

\input{tables/appendix_initialization_dfr}

\paragraph{Alternative initializations.}
We also evaluate alternative CXR initializations
(Table~\ref{tab:appendix_alternative_initialization}).
Using the M3AE-pretrained checkpoint provided by
CARZero~\cite{carzero} at the same input resolution
with a frozen copy of the initialized student as the DFR teacher,
\ours\ outperforms CARZero across all three task groups,
although grounding and segmentation at $224^2$ remain below
the best prior task-wise averages.
With a generic DINOv2~\cite{dinov2} initialization, \ours\ outperforms RadZero~\cite{radzero},
which uses radiology-adapted XrayDINOv2~\cite{chambon2024chexpertplus}, across all three task groups.
An ImageNet-21k pretrained Franca~\cite{venkataramanan2025franca} ViT-B/14 with a frozen copy of itself
as the teacher also exceeds the best prior task-wise averages
in all three groups, without medical pretraining or a larger teacher.

\input{tables/appendix_alternative_initialization}

\section{Variability across Random Seeds}

We report the mean and standard deviation over three independent trials
with different random seeds for \ours\ in zero-shot evaluation
and for all evaluated methods in downstream evaluation
(Tables~\ref{tab:appendix_cxr_zeroshot_stat_sig}--\ref{tab:appendix_ct_downstream_stat_sig}).

\input{tables/appendix_statistical_significance}

%% file: tables/appendix_training_details.tex
\begin{table}[t]
  \caption{
  Hyperparameters used for pretraining and downstream tasks.
  }
  \centering
  \normalsize
  \setlength{\tabcolsep}{5pt}
  \renewcommand{\arraystretch}{1.3}
  \resizebox{\columnwidth}{!}{
    \begin{NiceTabular}{cccccccc}
      \toprule
      
       & Task & Dataset & Batch size & Learning rate & Epoch
       & Early stopping
       & Loss \\
       
      \specialrule{0pt}{1pt}{0pt}
      
      \toprule
      
      \Block[c]{1-1}{Chest X-ray \\ training}  & VL Alignment & MIMIC-CXR  & 128 & 5e-5 & 10 & 3 & \ours  \\
      \midrule
     
      \Block[c]{3-1}{Chest X-ray \\ downstream }
     
      & \Block[c]{1-1}{Classification}
      & \Block[c]{1-1}{ChestXray14 \\ VinDr-CXR \\ RSNA-PN \\ SIIM} &  64 & 1e-4 & 20 & 3 &    BCE     \\
      \specialrule{0pt}{2pt}{3pt}
    
      & \Block[c]{1-1}{Report Generation} & MIMIC-CXR               &  64 & 1e-4 & 20 & 3 &    CE        \\
      \specialrule{0pt}{2pt}{2pt}
     
      & \Block[c]{1-1}{Segmentation}
      & \Block[c]{1-1}{ChestX-Det \\ SIIM}                          &  32 & 1e-4 & 40 & 3 & BCE, Dice  \\
      \specialrule{0pt}{0pt}{2pt}
     
     \toprule
     
     \Block[c]{1-1}{Chest CT \\ training}    & VL Alignment & CT-RATE       &  32  & 5e-5 & 5 & 3 & \ours   \\
     \midrule
     
     \Block[c]{4-1}{Chest CT \\ downstream}
     & \Block[c]{2-1}{Classification}  & CT-RATE                    & 128 & 5e-4 &  5 &  3 &    BCE     \\
     \specialrule{0pt}{0pt}{2pt}
     && \Block[c]{1-1} {RSNA-PE \\ LIDC-IDRI}                       &  32 & 5e-4 & 20 & 10 &    BCE      \\
     \specialrule{0pt}{1pt}{2pt}
     & \Block[c]{1-1}{Report Generation} & CT-RATE                  &  64 & 1e-4 &  5 &  3 & CE  \\
     \specialrule{0pt}{1pt}{2pt}
     & \Block[c]{1-1}{Segmentation}
      & \Block[c]{1-1}{MSD-Lung \\ NSCLC-Radiomics}                 &   8 & 5e-4 & 100 & 20 & CE, Dice  \\
     
    \bottomrule
        
      \CodeAfter
      \begin{tikzpicture}
          \begin{scope}[shorten <=2pt, shorten >=2pt]
            \draw (1-|2) -- (2-|2); 
            \draw (2-|2) -- (3-|2); 
            \draw (3-|2) -- (6-|2); 
            \draw (6-|2) -- (7-|2); 
            \draw (7-|2) -- (11-|2); 
            \draw (1-|3) -- (2-|3); 
            \draw (2-|3) -- (3-|3); 
            \draw (3-|3) -- (4-|3);
            \draw (4-|3) -- (5-|3);
            \draw (5-|3) -- (6-|3);
            \draw (6-|3) -- (7-|3);
            \draw (7-|3) -- (9-|3);
            \draw (9-|3) -- (10-|3);
            \draw (10-|3) -- (11-|3);
            \draw (1-|4) -- (2-|4); 
            \draw (2-|4) -- (3-|4); 
            \draw (3-|4) -- (4-|4);
            \draw (4-|4) -- (5-|4);
            \draw (5-|4) -- (6-|4);
            \draw (6-|4) -- (7-|4);
            \draw (7-|4) -- (9-|4);
            \draw (9-|4) -- (10-|4);
            \draw (10-|4) -- (11-|4);
        \end{scope}
        \begin{scope}[shorten <=2pt, shorten >=2pt]
            \draw (3-|2) -- (3-|9); 
            \draw (4-|2) -- (4-|9); 
            \draw (5-|2) -- (5-|9); 
            \draw (8-|3) -- (8-|9); 
            \draw (9-|2) -- (9-|9);
            \draw (10-|2) -- (10-|9);
        \end{scope}
      \end{tikzpicture}
    \end{NiceTabular}
  }
  \label{tab:appendix_training_details}
\end{table}

%% file: tables/computational_cost.tex
\begin{table}[t!]
\centering
\small
\caption{Computational cost of training.}
\begin{tabular}{lcccc}
\toprule
Model & GPUs & Memory / GPU & Total Time & GPU-hours \\
\midrule
\ours-CXR & 8 $\times$ H200 & 128.6\,GB & 5.5 h & 44.0 \\
\ours-CT  & 8 $\times$ H200 & 148.0\,GB & 4.3 h & 34.4 \\
\bottomrule
\end{tabular}
\label{tab:compute_cost}
\end{table}

%% file: tables/appendix_test_dataset.tex
\begin{table}[t]
    \caption{Evaluation dataset details, including the number of instances per split and key annotation properties, such as class count and annotation type. 
    }
  \centering
  \small
  \setlength{\tabcolsep}{5pt}
  \renewcommand{\arraystretch}{1.15}

  \resizebox{\columnwidth}{!}{
    \begin{NiceTabular}{ccccccc}
      \toprule
      
       & Task & Dataset & Train & Validation & Test & Annotation \\
      \specialrule{0pt}{1pt}{0pt}
      
      \toprule
      
      \Block[c]{10-1}{Chest X-ray \\ zero-shot}
      & \Block[c]{6-1}{Classification}
      & PadChest       &-&-& 39,053 &  192 classes \\
      & & PadChest20     &-&-& 39,053 &   20 classes \\
      & & ChestXray14    &-&-& 22,433 &   14 classes \\
      & & ChestX-Det10   &-&-&    542 & 10 classes, bounding box \\
      & & CheXpert       &-&-&    668 &    5 classes \\
      & & Open-I         &-&-&  3,851 &   18 classes \\
      \specialrule{0pt}{2pt}{2pt}
      & \Block[c]{2-1}{Grounding}
      & ChestX-Det10   &-&-&    542 & 10 classes, bounding box \\
      && MS-CXR         &-&-&    167 & phrase-level bounding box \\
      \specialrule{0pt}{1pt}{2pt}
      & \Block[c]{2-1}{Segmentation}
      & ChestX-Det     &-&-&  3,578 & 13 classes, mask \\
      & & SIIM           &-&-&  1,704 &  1 class, mask       \\
    
     \midrule
     
     \Block[c]{7-1}{Chest X-ray \\ downstream }
     & \Block[c]{4-1}{Classification}
     & ChestXray14    & 77,871 & 8,653 & 25,596 &  14 classes  \\
     & & VinDr-CXR      & 13,494 & 1,506 &  3,000 &  28 classes, bounding box \\
     & & RSNA-PN        & 21,346 & 2,669 &  2,669 &  1 class     \\
     & & SIIM           &  9,303 & 1,372 &  1,372 &  1 class, mask \\
     \specialrule{0pt}{2pt}{2pt}
     & \Block[c]{1-1}{Report Generation}
     & MIMIC-CXR      & 27,079 & 2,130 &  3,858 &   report     \\
     \specialrule{0pt}{2pt}{2pt}
     & \Block[c]{2-1}{Segmentation}
     & ChestX-Det     &  2,722 &   303 &    553 & 13 classes, mask \\
     & & SIIM           &  9,303 & 1,372 &  1,372 &  1 class, mask \\
     \specialrule{0pt}{0pt}{2pt}
     
     \toprule
     
     \Block[c]{3-1}{Chest CT \\ zero-shot}
     & \Block[c]{2-1}{Classification}
     & CT-RATE        &-&-& 3,002 & 18 classes, report  \\
     & & RAD-ChestCT    &-&-& 3,630 & 16 classes \\
     \specialrule{0pt}{1pt}{2pt}
     & \Block[c]{1-1}{Free-text Segmentation}
     & ReXGroundingCT &-&-&    50 & free-text, mask \\

     \midrule
     
     \Block[c]{6-1}{Chest CT \\ downstream}
     & \Block[c]{3-1}{Classification}
     & CT-RATE        & 43,390 & 2,993 & 3,002 & 18 classes, report \\
     & & RSNA-PE        &  4,983 & 1,062 & 1,060 & 1 class  \\
     & & LIDC-IDRI      &    705 &   148 &   157 & 1 class  \\
     \specialrule{0pt}{1pt}{2pt}
     & \Block[c]{1-1}{Report Generation}
     & CT-RATE        & 43,390 & 2,993 & 3,002 & 18 classes, report \\
     \specialrule{0pt}{1pt}{2pt}
     & \Block[c]{2-1}{Segmentation}
     & MSD-Lung        &     44 &     6 &    13 & 1 class, mask  \\
     & & NSCLC-Radiomics &    294 &    42 &    84 & 1 class, mask \\
     
     \bottomrule
        
      \CodeAfter
      \begin{tikzpicture}
          \begin{scope}[shorten <=2pt, shorten >=2pt]
            \draw (1-|2) -- (2-|2);
            \draw (2-|2) -- (12-|2);
            \draw (12-|2) -- (19-|2);
            \draw (19-|2) -- (22-|2);
            \draw (22-|2) -- (28-|2);

            \draw (1-|3) -- (2-|3);
            \draw (2-|3) -- (8-|3);
            \draw (8-|3) -- (10-|3);
            \draw (10-|3) -- (12-|3);
            \draw (12-|3) -- (16-|3);
            \draw (16-|3) -- (17-|3);
            \draw (17-|3) -- (19-|3);
            \draw (19-|3) -- (21-|3);
            \draw (21-|3) -- (22-|3);
            \draw (22-|3) -- (25-|3);
            \draw (25-|3) -- (26-|3);
            \draw (26-|3) -- (28-|3);

            \draw (1-|4) -- (2-|4);
            \draw (2-|4) -- (8-|4);
            \draw (8-|4) -- (10-|4);
            \draw (10-|4) -- (12-|4);
            \draw (12-|4) -- (16-|4);
            \draw (16-|4) -- (17-|4);
            \draw (17-|4) -- (19-|4);
            \draw (19-|4) -- (21-|4);
            \draw (21-|4) -- (22-|4);
            \draw (22-|4) -- (25-|4);
            \draw (25-|4) -- (26-|4);
            \draw (26-|4) -- (28-|4);

            \draw (1-|7) -- (2-|7);
            \draw (2-|7) -- (8-|7);
            \draw (8-|7) -- (10-|7);
            \draw (10-|7) -- (12-|7);
            \draw (12-|7) -- (16-|7);
            \draw (16-|7) -- (17-|7);
            \draw (17-|7) -- (19-|7);
            \draw (19-|7) -- (21-|7);
            \draw (21-|7) -- (22-|7);
            \draw (22-|7) -- (25-|7);
            \draw (25-|7) -- (26-|7);
            \draw (26-|7) -- (28-|7);

        \end{scope}
        \begin{scope}[shorten <=2pt, shorten >=2pt]
            \draw (8-|2) -- (8-|8);
            \draw (10-|2) -- (10-|8);
            \draw (16-|2) -- (16-|8);
            \draw (17-|2) -- (17-|8);
            \draw (21-|2) -- (21-|8);
            \draw (25-|2) -- (25-|8);
            \draw (26-|2) -- (26-|8);

        \end{scope}
      \end{tikzpicture}
    \end{NiceTabular}
  } 
  \label{tab:appendix_test_dataset}
\end{table}

%% file: tables/appendix_dfr_ablation.tex
\begin{table*}[!t]
  \caption{
    Ablation on DFR. ViT-7B/G: ViT-7B (CXR), ViT-G (CT). \ours\ is highlighted in gray.
  }
  \centering
  \small
  \setlength{\tabcolsep}{5pt}
  \renewcommand{\arraystretch}{1.15}

  \resizebox{\textwidth}{!}{
    \begin{NiceTabular}{l cc *{3}{wc{1.55cm}} *{2}{wc{1.55cm}}}[color-inside]
      \toprule

      \Block[c]{2-1}{\textbf{Variant}}
        & \Block[c]{2-1}{\textbf{Teacher}}
        & \Block[c]{2-1}{\textbf{Depth}}
        & \multicolumn{3}{c}{\textbf{CXR Zero-shot}}
        & \multicolumn{2}{c}{\textbf{Chest CT Zero-shot}} \\
      \specialrule{0pt}{3pt}{3pt}

        & &
        & Classification & Grounding & Segmentation
        & Classification & Segmentation \\
      \specialrule{0pt}{2pt}{0pt}
      \toprule

      \multicolumn{8}{l}{\textit{(a) Alternative regularization}} \\
      \quad Freeze all but last-2 student layers (no DFR)
          & --        & --
          & 0.855 & 0.750 & 0.281 & 0.798 & 0.097 \\

      \midrule
      \multicolumn{8}{l}{\textit{(b) Teacher capacity \,(DFR, depth fixed)}} \\
      \quad Same-size teacher
          & ViT-L     & 90\%
          & 0.863 & 0.773 & 0.335 & 0.811 & 0.127 \\

      \midrule
      \multicolumn{8}{l}{\textit{(c) Regularization depth \,(DFR, ViT-7B/G teacher)}} \\
      \quad Depth 50\%
          & ViT-7B/G  & 50\%
          & 0.875 & 0.787 & 0.352 & 0.800 & 0.114 \\
      \quad Depth 75\%
          & ViT-7B/G  & 75\%
          & 0.870 & 0.785 & 0.356 & 0.791 & 0.094 \\
      \quad Depth 100\%
          & ViT-7B/G  & 100\%
          & 0.871 & \textbf{0.826} & 0.303 & 0.811 & 0.100 \\

      \specialrule{0pt}{0pt}{1pt}
      \rowcolor{gray!20}
      \quad \ours\ 
          & ViT-7B/G  & 90\%
          & \textbf{0.877} & 0.810 & \textbf{0.375} & \textbf{0.819} & \textbf{0.128} \\
      \specialrule{0.7pt}{0pt}{-1pt}

      \CodeAfter
      \begin{tikzpicture}
        \begin{scope}[shorten <=2pt, shorten >=2pt]
          \draw (1-|4) -- (3-|4);
          \draw (1-|7) -- (3-|7);
        \end{scope}
        \begin{scope}[shorten <=3pt, shorten >=3pt]
          \draw (2-|4) -- (2-|7);    
          \draw (2-|7) -- (2-|9);    
        \end{scope}
        \begin{scope}[shorten <=2pt, shorten >=2pt]
          \draw (3-|4) -- (5-|4);
          \draw (3-|7) -- (5-|7);
        \end{scope}
        \begin{scope}[shorten <=2pt, shorten >=2pt]
          \draw (5-|4) -- (7-|4);
          \draw (5-|7) -- (7-|7);
        \end{scope}
        \begin{scope}[shorten <=2pt, shorten >=2pt]
          \draw (7-|4) -- (last-|4);
          \draw (7-|7) -- (last-|7);
        \end{scope}
      \end{tikzpicture}

    \end{NiceTabular}
  }
  \label{tab:appendix_dfr_ablation}
\end{table*}

%% file: tables/appendix_cxr_resolution_ablation.tex
\begin{table*}[!t]
  \caption{
    Ablation on input resolution and vision encoder size for chest X-ray. \ours-CXR is highlighted in gray.
  }
  \centering
  \small
  \resizebox{\textwidth}{!}{
    \begin{NiceTabular}{cc *{6}{wc{1.4cm}} *{2}{wc{1.6cm}} *{2}{wc{1.6cm}}}[color-inside]
      \toprule

      \Block[c]{3-1}{\textbf{Resolution}}
        & \Block[c]{3-1}{\textbf{Encoder}}
        & \multicolumn{6}{c}{\textbf{Classification}}
        & \multicolumn{2}{c}{\textbf{Grounding}}
        & \multicolumn{2}{c}{\textbf{Segmentation}} \\
      \specialrule{0pt}{2pt}{3pt}

        &
        & PadChest & PadChest20 & ChestXray14 & ChestX-Det10 & CheXpert & Open-I
        & ChestX-Det10 & MS-CXR
        & ChestX-Det & SIIM \\
      \specialrule{0pt}{2pt}{3pt}

        &
        & \multicolumn{6}{c}{Macro AUROC}
        & Macro Acc.
        & Acc.
        & Macro Dice
        & Dice \\
      \specialrule{0pt}{2pt}{0pt}
      \toprule

      $512^2$  & ViT-B
        & 0.853 & 0.894 & 0.817 & 0.812 & 0.918 & 0.871
        & 0.625 & 0.886
        & 0.385 & 0.259 \\

      $1024^2$ & ViT-B
        & 0.853 & 0.891 & \textbf{0.826} & 0.838 & 0.913 & 0.880
        & 0.681 & \textbf{0.916}
        & 0.419 & \textbf{0.317} \\

      $512^2$  & ViT-L
        & 0.853 & 0.885 & 0.823 & 0.831 & 0.911 & 0.870
        & 0.628 & 0.880
        & 0.408 & 0.224 \\

      \specialrule{0pt}{0pt}{1pt}
      \rowcolor{gray!20}
      $1024^2$ & ViT-L
        & \textbf{0.863} & \textbf{0.917} & 0.825 & \textbf{0.853} & \textbf{0.921} & \textbf{0.881}
        & \textbf{0.729} & 0.892
        & \textbf{0.434} & \textbf{0.317} \\
      \specialrule{0.7pt}{0pt}{-1pt}

      \CodeAfter
      \begin{tikzpicture}
        \begin{scope}[shorten <=2pt, shorten >=2pt]
          \draw (1-|3) -- (4-|3);
          \draw (1-|9) -- (4-|9);
          \draw (1-|11) -- (4-|11);
          \draw (4-|3) -- (8-|3);
          \draw (4-|9) -- (8-|9);
          \draw (4-|11) -- (8-|11);
        \end{scope}
        \begin{scope}[shorten <=3pt, shorten >=3pt]
          \draw (2-|3) -- (2-|9);   
          \draw (2-|9) -- (2-|11);  
          \draw (2-|11) -- (2-|13); 
          \draw (3-|3) -- (3-|9);
          \draw (3-|9) -- (3-|10);
          \draw (3-|10) -- (3-|11);
          \draw (3-|11) -- (3-|12);
          \draw (3-|12) -- (3-|13);
        \end{scope}
      \end{tikzpicture}

    \end{NiceTabular}
  }
  \label{tab:appendix_cxr_resolution_ablation}
\end{table*}

%% file: tables/appendix_initialization_dfr.tex
\begin{table*}[t]
  \caption{Effects of pretrained initialization and DFR on zero-shot
  performance. Results are task-wise averages.
  Bold denotes the highest value in each column.}
  \label{tab:appendix_initialization_dfr}
  \centering
  \small
  \setlength{\tabcolsep}{5pt}
  \renewcommand{\arraystretch}{1.15}
  \resizebox{\textwidth}{!}{%
    \begin{NiceTabular}{lc*{5}{c}}[color-inside]
      \toprule
      Initialization & DFR
      & \multicolumn{3}{c}{\textbf{CXR Zero-shot}}
      & \multicolumn{2}{c}{\textbf{Chest CT Zero-shot}} \\
      \cmidrule(lr){3-5}
      \cmidrule(lr){6-7}
      & & Classification & Grounding & Segmentation
        & Classification & Segmentation \\
      \midrule
      Best prior work & --
      & 0.848 & 0.733 & 0.252 & 0.760 & -- \\
      \midrule
      \rowcolor{gray!20}
      Pretrained & \ding{51}
      & \textbf{0.877} & \textbf{0.810} & \textbf{0.375}
      & \textbf{0.819} & \textbf{0.128} \\
      Pretrained & --
      & 0.870 & 0.772 & 0.369 & 0.812 & 0.119 \\
      From scratch & \ding{51}
      & 0.822 & 0.677 & 0.195 & 0.781 & 0.114 \\
      From scratch & --
      & 0.776 & 0.535 & 0.111 & 0.736 & 0.005 \\
      \bottomrule
    \end{NiceTabular}%
  }
\end{table*}

%% file: tables/appendix_alternative_initialization.tex
\begin{table*}[t]
  \caption{CXR zero-shot performance with alternative initializations.
  Results are task-wise averages.
  Bold denotes scores above the best prior result in each task group.
  ``Self'' denotes a frozen copy of the initialized student.}
  \label{tab:appendix_alternative_initialization}
  \centering
  \small
  \setlength{\tabcolsep}{4pt}
  \renewcommand{\arraystretch}{1.15}
  \resizebox{\textwidth}{!}{%
    \begin{NiceTabular}{llccccc}
      \toprule
      Method & Initialization & Resolution
      & \makecell{DFR\\teacher}
      & Classification & Grounding & Segmentation \\
      \midrule
      Best prior work & -- & -- & --
      & 0.848 & 0.733 & 0.252 \\
      \midrule
      CARZero~\cite{carzero}
       & \makecell[l]{ViT-B/16\\M3AE (CARZero)~\cite{carzero}}
      & $224^2$ & --
      & 0.836 & 0.646 & 0.193 \\
      \ours
      & \makecell[l]{ViT-B/16\\M3AE (CARZero)~\cite{carzero}}
      & $224^2$ & Self
      & \textbf{0.856} & 0.671 & 0.229 \\
      \addlinespace[3pt]
      RadZero~\cite{radzero}
      & \makecell[l]{ViT-B/14\\
        XrayDINOv2~\cite{chambon2024chexpertplus}}
      & $518^2$ & --
      & 0.842 & 0.733 & 0.252 \\
      \ours
      & \makecell[l]{ViT-B/14\\DINOv2~\cite{dinov2}}
      & $518^2$ & ViT-g/14
      & \textbf{0.852} & \textbf{0.762} & \textbf{0.327} \\
      \addlinespace[3pt]
      \ours
      & \makecell[l]{ViT-B/14\\
        Franca on ImageNet-21k~\cite{venkataramanan2025franca}}
      & $518^2$ & Self
      & \textbf{0.857} & \textbf{0.760} & \textbf{0.300} \\
      \bottomrule
    \end{NiceTabular}%
  }
\end{table*}

%% file: tables/appendix_statistical_significance.tex
\noindent\begin{minipage}{\textwidth}
\centering
\captionsetup{type=table}
\captionof{table}{Zero-shot performance of \ours-CXR on chest X-ray (mean $\pm$ standard deviation over three random seeds).}
\label{tab:appendix_cxr_zeroshot_stat_sig}
\small
\setlength{\tabcolsep}{4pt}
\renewcommand{\arraystretch}{1.15}
\resizebox{\textwidth}{!}{%
  \begin{NiceTabular}{p{2.2cm} *{6}{wc{1.5cm}} *{2}{wc{1.85cm}} *{2}{wc{1.65cm}}}
    \toprule
      \Block[c]{3-1}{Method}
        & \multicolumn{6}{c}{\textbf{Classification}}
        & \multicolumn{2}{c}{\textbf{Grounding}}
        & \multicolumn{2}{c}{\textbf{Segmentation}} \\
      \specialrule{0pt}{2pt}{3pt}

        & PadChest & PadChest20 & ChestXray14 & ChestX-Det10 & CheXpert & Open-I
        & ChestX-Det10 & MS-CXR
        & ChestX-Det & SIIM \\
      \specialrule{0pt}{2pt}{3pt}

      & \multicolumn{6}{c}{Macro AUROC}
      & Macro Acc.
      & Acc.
      & Macro Dice
      & Dice \\
      \specialrule{0pt}{2pt}{0pt}
      \toprule

      \multicolumn{1}{c}{\textbf{\ours-CXR}}
        & 0.863 & 0.917 & 0.825 & 0.853 & 0.921 & 0.881
        & 0.729 & 0.892
        & 0.434 & 0.317 \\

      \specialrule{0pt}{0pt}{1pt}
      \makecell{Mean ($\pm$ std)}
          & \makecell{0.860 \\ ($\pm$ 0.0026)}
          & \makecell{0.908 \\ ($\pm$ 0.0090)}
          & \makecell{0.827 \\ ($\pm$ 0.0028)}
          & \makecell{0.845 \\ ($\pm$ 0.0073)}
          & \makecell{0.915 \\ ($\pm$ 0.0056)}
          & \makecell{0.879 \\ ($\pm$ 0.0024)}
          & \makecell{0.711 \\ ($\pm$ 0.0222)}
          & \makecell{0.904 \\ ($\pm$ 0.0317)}
          & \makecell{0.429 \\ ($\pm$ 0.0044)}
          & \makecell{0.305 \\ ($\pm$ 0.0246)} \\
      \specialrule{0.7pt}{0pt}{-1pt}

      \CodeAfter
      \begin{tikzpicture}
          \begin{scope}[shorten <=2pt, shorten >=2pt]
            \draw (1-|2) -- (4-|2);
            \draw (1-|8) -- (4-|8);
            \draw (1-|10) -- (4-|10);
            \draw (4-|2) -- (6-|2);
            \draw (4-|8) -- (6-|8);
            \draw (4-|10) -- (6-|10);
        \end{scope}
        \begin{scope}[shorten <=3pt, shorten >=3pt]
            \draw (2-|2) -- (2-|8);   
            \draw (2-|8) -- (2-|10);  
            \draw (2-|10) -- (2-|12); 
            \draw (3-|2) -- (3-|8);
            \draw (3-|8) -- (3-|9);
            \draw (3-|9) -- (3-|10);
            \draw (3-|10) -- (3-|11);
            \draw (3-|11) -- (3-|12);
        \end{scope}
      \end{tikzpicture}
  \end{NiceTabular}%
}
\end{minipage}

\par\vspace{18pt}

\noindent\begin{minipage}{\textwidth}
  \captionof{table}{Downstream task performance on chest X-ray (mean $\pm$ standard deviation over three random seeds). Bold denotes the highest mean performance.}
  \centering
  \small
  \setlength{\tabcolsep}{5pt}
  \renewcommand{\arraystretch}{1.15}

  \resizebox{\textwidth}{!}{
    \begin{NiceTabular}{l *{4}{wc{1.6cm}} *{4}{wc{1.6cm}} *{2}{wc{1.6cm}}}[color-inside]
      \toprule
      \Block[c]{3-1}{Method}

      & \multicolumn{4}{c}{\textbf{Classification}}
      & \multicolumn{4}{c}{\textbf{Report Generation}}
      & \multicolumn{2}{c}{\textbf{Segmentation}} \\
      \specialrule{0pt}{2pt}{3pt}

      & ChestXray14 & VinDr-CXR & RSNA-PN & SIIM
      & \multicolumn{4}{c}{MIMIC-CXR}
      & ChestX-Det & SIIM \\
      \specialrule{0pt}{2pt}{3pt}

      & \multicolumn{2}{c}{Macro AUROC}
      & \multicolumn{2}{c}{AUROC}
      & METEOR & ROUGE-L & CheXbert F1 & F1-RadGraph
      & Macro Dice & Dice \\
      \specialrule{0pt}{2pt}{0pt}
      \toprule

      DINOv3~\cite{dinov3}
      & \makecell{0.806 \\ ($\pm$ 0.001)}
      & \makecell{0.883 \\ ($\pm$ 0.015)}
      & \makecell{0.879 \\ ($\pm$ 0.003)}
      & \makecell{0.931 \\ ($\pm$ 0.006)}
      & \makecell{0.130 \\ ($\pm$ 0.003)}
      & \makecell{0.238 \\ ($\pm$ 0.002)}
      & \makecell{0.238 \\ ($\pm$ 0.004)}
      & \makecell{0.202 \\ ($\pm$ 0.004)}
      & \makecell{0.765 \\ ($\pm$ 0.065)}
      & \makecell{0.479 \\ ($\pm$ 0.077)} \\
      \addlinespace[4pt]

      MGCA~\cite{mgca}
      & \makecell{0.808 \\ ($\pm$ 0.001)}
      & \makecell{0.923 \\ ($\pm$ 0.000)}
      & \makecell{0.892 \\ ($\pm$ 0.001)}
      & \makecell{0.928 \\ ($\pm$ 0.003)}
      & \makecell{0.131 \\ ($\pm$ 0.001)}
      & \makecell{0.237 \\ ($\pm$ 0.001)}
      & \makecell{0.259 \\ ($\pm$ 0.008)}
      & \makecell{0.204 \\ ($\pm$ 0.002)}
      & \makecell{0.779 \\ ($\pm$ 0.041)}
      & \makecell{0.514 \\ ($\pm$ 0.110)} \\
      \addlinespace[4pt]

      KAD~\cite{kad}
      & \makecell{0.812 \\ ($\pm$ 0.003)}
      & \makecell{0.934 \\ ($\pm$ 0.001)}
      & \makecell{0.896 \\ ($\pm$ 0.002)}
      & \makecell{0.937 \\ ($\pm$ 0.002)}
      & \makecell{0.135 \\ ($\pm$ 0.002)}
      & \makecell{0.243 \\ ($\pm$ 0.001)}
      & \makecell{0.277 \\ ($\pm$ 0.005)}
      & \makecell{0.215 \\ ($\pm$ 0.003)}
      & \makecell{0.789 \\ ($\pm$ 0.007)}
      & \makecell{0.539 \\ ($\pm$ 0.021)} \\
      \addlinespace[4pt]

      MedKLIP~\cite{medklip}
      & \makecell{0.801 \\ ($\pm$ 0.000)}
      & \makecell{0.919 \\ ($\pm$ 0.005)}
      & \makecell{0.893 \\ ($\pm$ 0.001)}
      & \makecell{0.937 \\ ($\pm$ 0.002)}
      & \makecell{0.135 \\ ($\pm$ 0.001)}
      & \makecell{0.237 \\ ($\pm$ 0.001)}
      & \makecell{0.268 \\ ($\pm$ 0.007)}
      & \makecell{0.204 \\ ($\pm$ 0.002)}
      & \makecell{0.781 \\ ($\pm$ 0.020)}
      & \makecell{0.507 \\ ($\pm$ 0.009)} \\
      \addlinespace[4pt]

      MRM~\cite{mrm}
      & \makecell{0.818 \\ ($\pm$ 0.000)}
      & \makecell{0.907 \\ ($\pm$ 0.006)}
      & \makecell{0.893 \\ ($\pm$ 0.001)}
      & \makecell{0.937 \\ ($\pm$ 0.000)}
      & \makecell{0.133 \\ ($\pm$ 0.003)}
      & \makecell{0.240 \\ ($\pm$ 0.004)}
      & \makecell{0.271 \\ ($\pm$ 0.003)}
      & \makecell{0.209 \\ ($\pm$ 0.002)}
      & \makecell{0.793 \\ ($\pm$ 0.014)}
      & \makecell{0.577 \\ ($\pm$ 0.048)} \\
      \addlinespace[4pt]

      MAVL~\cite{mavl}
      & \makecell{0.802 \\ ($\pm$ 0.001)}
      & \makecell{0.914 \\ ($\pm$ 0.004)}
      & \makecell{0.892 \\ ($\pm$ 0.002)}
      & \makecell{0.944 \\ ($\pm$ 0.001)}
      & \makecell{0.133 \\ ($\pm$ 0.001)}
      & \makecell{0.240 \\ ($\pm$ 0.002)}
      & \makecell{0.278 \\ ($\pm$ 0.005)}
      & \makecell{0.211 \\ ($\pm$ 0.003)}
      & \makecell{0.765 \\ ($\pm$ 0.003)}
      & \makecell{0.551 \\ ($\pm$ 0.037)} \\
      \addlinespace[4pt]

      CheXWorld~\cite{chexworld}
      & \makecell{0.814 \\ ($\pm$ 0.002)}
      & \makecell{0.922 \\ ($\pm$ 0.004)}
      & \makecell{0.888 \\ ($\pm$ 0.003)}
      & \makecell{0.937 \\ ($\pm$ 0.002)}
      & \makecell{0.133 \\ ($\pm$ 0.000)}
      & \makecell{0.238 \\ ($\pm$ 0.001)}
      & \makecell{0.259 \\ ($\pm$ 0.006)}
      & \makecell{0.208 \\ ($\pm$ 0.001)}
      & \makecell{0.795 \\ ($\pm$ 0.006)}
      & \makecell{0.586 \\ ($\pm$ 0.007)} \\
      \addlinespace[4pt]

      RAD-DINO~\cite{raddino}
      & \makecell{0.831 \\ ($\pm$ 0.001)}
      & \makecell{0.936 \\ ($\pm$ 0.003)}
      & \makecell{0.891 \\ ($\pm$ 0.001)}
      & \makecell{0.953 \\ ($\pm$ 0.002)}
      & \makecell{0.137 \\ ($\pm$ 0.001)}
      & \makecell{0.242 \\ ($\pm$ 0.001)}
      & \makecell{0.283 \\ ($\pm$ 0.007)}
      & \makecell{0.213 \\ ($\pm$ 0.000)}
      & \makecell{0.808 \\ ($\pm$ 0.003)}
      & \makecell{0.631 \\ ($\pm$ 0.054)} \\
      \addlinespace[4pt]

      \specialrule{0pt}{0pt}{1pt}
      \rowcolor{gray!20}
      \ours-CXR
      & \makecell{\textbf{0.847} \\ ($\pm$ 0.001)}
      & \makecell{\textbf{0.944} \\ ($\pm$ 0.001)}
      & \makecell{\textbf{0.902} \\ ($\pm$ 0.001)}
      & \makecell{\textbf{0.975} \\ ($\pm$ 0.001)}
      & \makecell{\textbf{0.143} \\ ($\pm$ 0.001)}
      & \makecell{\textbf{0.249} \\ ($\pm$ 0.002)}
      & \makecell{\textbf{0.314} \\ ($\pm$ 0.002)}
      & \makecell{\textbf{0.225} \\ ($\pm$ 0.002)}
      & \makecell{\textbf{0.818} \\ ($\pm$ 0.007)}
      & \makecell{\textbf{0.730} \\ ($\pm$ 0.049)} \\
      \specialrule{0.7pt}{0pt}{-1pt}

      \CodeAfter
      \begin{tikzpicture}
          \begin{scope}[shorten <=2pt, shorten >=2pt]
            \draw (1-|2) -- (4-|2);
            \draw (1-|6) -- (4-|6);
            \draw (1-|10) -- (4-|10);
            \draw (4-|2) -- (13-|2);
            \draw (4-|6) -- (13-|6);
            \draw (4-|10) -- (13-|10);
        \end{scope}
        \begin{scope}[shorten <=3pt, shorten >=3pt]
            \draw (2-|2) -- (2-|6);   
            \draw (2-|6) -- (2-|10);  
            \draw (2-|10) -- (2-|12); 
            \draw (3-|2) -- (3-|4);
            \draw (3-|4) -- (3-|6);
            \draw (3-|6) -- (3-|10);
            \draw (3-|10) -- (3-|11);
            \draw (3-|11) -- (3-|12);

        \end{scope}
      \end{tikzpicture}
    \end{NiceTabular}
  } 
  \label{tab:appendix_cxr_downstream_stat_sig}
\end{minipage}

\par\vspace{18pt}

\noindent\begin{minipage}{\textwidth}
\centering
\captionsetup{type=table}
\captionof{table}{Zero-shot performance of \ours-CT on chest CT (mean $\pm$ standard deviation over three random seeds).}
\label{tab:appendix_ct_zeroshot_stat_sig}
\small
\setlength{\tabcolsep}{5pt}
\renewcommand{\arraystretch}{1.15}
\resizebox{0.5\textwidth}{!}{%
  \begin{NiceTabular}{p{2.2cm} *{2}{wc{2.0cm}} *{1}{wc{2.6cm}}}
      \toprule
      \Block[c]{3-1}{Method}
        & \multicolumn{2}{c}{\textbf{Classification}}
        & \multicolumn{1}{c}{\textbf{Free-text Segmentation}} \\
        \specialrule{0pt}{2pt}{2pt}

        & CT-RATE & RAD-ChestCT
        & ReXGroundingCT \\
        \specialrule{0pt}{2pt}{2pt}

        & \multicolumn{2}{c}{Macro AUROC}
        & Dice \\
        \specialrule{0pt}{1pt}{0pt}
        \toprule

      \multicolumn{1}{c}{\textbf{\ours-CT}}
        & 0.850 & 0.789
        & 0.128 \\

      \specialrule{0pt}{0pt}{2pt}
      \makecell{Mean ($\pm$ std)}
        & \makecell{0.849 \\ ($\pm$ 0.0010)}
        & \makecell{0.780 \\ ($\pm$ 0.0098)}
        & \makecell{0.122 \\ ($\pm$ 0.0098)} \\
      \specialrule{0pt}{0pt}{2pt}
      \bottomrule

      \CodeAfter
      \begin{tikzpicture}
          \begin{scope}[shorten <=2pt, shorten >=2pt]
            \draw (1-|2) -- (4-|2);
            \draw (1-|4) -- (4-|4);
            \draw (4-|2) -- (6-|2);
            \draw (4-|4) -- (6-|4);
        \end{scope}
        \begin{scope}[shorten <=2pt, shorten >=2pt]
            \draw (2-|2) -- (2-|4);   
            \draw (2-|4) -- (2-|5);   
            \draw (3-|2) -- (3-|4);
            \draw (3-|4) -- (3-|5);
        \end{scope}
      \end{tikzpicture}
  \end{NiceTabular}%
}
\end{minipage}

\par\vspace{18pt}

\noindent\begin{minipage}{\textwidth}
\centering
\small
\captionof{table}{Downstream task performance on chest CT (mean $\pm$ standard deviation over three random seeds). Bold denotes the highest mean performance.}
  \centering
  \small
  \setlength{\tabcolsep}{2pt}
  \renewcommand{\arraystretch}{1.15}
  \resizebox{\textwidth}{!}{%
    \begin{NiceTabular}{p{2.2cm} *{3}{wc{2.0cm}} *{4}{wc{1.8cm}} *{2}{wc{1.7cm}}}[color-inside]
      \toprule
      \Block[c]{3-1}{Method}
      & \multicolumn{3}{c}{\textbf{Classification}}
      & \multicolumn{4}{c}{\textbf{Report Generation}}
      & \multicolumn{2}{c}{\textbf{Segmentation}} \\
      \specialrule{0pt}{2pt}{3pt}
      & CT-RATE & RSNA-PE & LIDC-IDRI
      & \multicolumn{4}{c}{CT-RATE}
      & MSD-Lung & NSCLC-Rad \\
      \specialrule{0pt}{2pt}{3pt}
      & Macro AUROC
      & \multicolumn{2}{c}{AUROC}
      & METEOR & ROUGE-L & RadBERT F1 & CRG
      & \multicolumn{2}{c}{Dice} \\
      \specialrule{0pt}{2pt}{0pt}
      \toprule
      V-JEPA 2.1~\cite{vjepa21}
      & \makecell{0.857 \\ ($\pm$ 0.003)}
      & \makecell{0.627 \\ ($\pm$ 0.015)}
      & \makecell{0.788 \\ ($\pm$ 0.017)}
      & \makecell{0.223 \\ ($\pm$ 0.008)}
      & \makecell{0.271 \\ ($\pm$ 0.018)}
      & \makecell{0.338 \\ ($\pm$ 0.007)}
      & \makecell{0.397 \\ ($\pm$ 0.004)}
      & \makecell{0.532 \\ ($\pm$ 0.037)}
      & \makecell{0.561 \\ ($\pm$ 0.003)} \\
      \addlinespace[4pt]
      CT-CLIP~\cite{ctclip}
      & \makecell{0.766 \\ ($\pm$ 0.002)}
      & \makecell{0.595 \\ ($\pm$ 0.019)}
      & \makecell{0.786 \\ ($\pm$ 0.008)}
      & \makecell{0.191 \\ ($\pm$ 0.005)}
      & \makecell{0.274 \\ ($\pm$ 0.005)}
      & \makecell{0.124 \\ ($\pm$ 0.003)}
      & \makecell{0.352 \\ ($\pm$ 0.002)}
      & \makecell{0.022 \\ ($\pm$ 0.035)}
      & \makecell{0.360 \\ ($\pm$ 0.017)} \\
      \addlinespace[4pt]
      fVLM~\cite{fvlm}
      & \makecell{0.807 \\ ($\pm$ 0.000)}
      & \makecell{0.649 \\ ($\pm$ 0.009)}
      & \makecell{0.730 \\ ($\pm$ 0.029)}
      & \makecell{0.224 \\ ($\pm$ 0.001)}
      & \makecell{0.290 \\ ($\pm$ 0.001)}
      & \makecell{0.327 \\ ($\pm$ 0.007)}
      & \makecell{0.394 \\ ($\pm$ 0.003)}
      & --
      & -- \\
      \addlinespace[4pt]
      ViSD-Boost~\cite{visd-boost}
      & \makecell{0.801 \\ ($\pm$ 0.001)}
      & \makecell{0.601 \\ ($\pm$ 0.002)}
      & \makecell{0.827 \\ ($\pm$ 0.004)}
      & \makecell{0.219 \\ ($\pm$ 0.001)}
      & \makecell{\textbf{0.294} \\ ($\pm$ 0.000)}
      & \makecell{0.293 \\ ($\pm$ 0.007)}
      & \makecell{0.384 \\ ($\pm$ 0.003)}
      & --
      & -- \\
      \addlinespace[4pt]
      COLIPRI~\cite{colipri}
      & \makecell{\textbf{0.876} \\ ($\pm$ 0.000)}
      & \makecell{0.673 \\ ($\pm$ 0.009)}
      & \makecell{0.874 \\ ($\pm$ 0.024)}
      & \makecell{\textbf{0.228} \\ ($\pm$ 0.009)}
      & \makecell{0.289 \\ ($\pm$ 0.004)}
      & \makecell{0.391 \\ ($\pm$ 0.018)}
      & \makecell{0.425 \\ ($\pm$ 0.008)}
      & \makecell{0.568 \\ ($\pm$ 0.052)}
      & \makecell{0.581 \\ ($\pm$ 0.012)} \\
      \addlinespace[4pt]
      \specialrule{0pt}{0pt}{1pt}
      \rowcolor{gray!20}
      \ours-CT
      & \makecell{\textbf{0.876} \\ ($\pm$ 0.002)}
      & \makecell{\textbf{0.677} \\ ($\pm$ 0.021)}
      & \makecell{\textbf{0.901} \\ ($\pm$ 0.015)}
      & \makecell{0.221 \\ ($\pm$ 0.005)}
      & \makecell{0.226 \\ ($\pm$ 0.017)}
      & \makecell{\textbf{0.399} \\ ($\pm$ 0.001)}
      & \makecell{\textbf{0.432} \\ ($\pm$ 0.003)}
      & \makecell{\textbf{0.613} \\ ($\pm$ 0.031)}
      & \makecell{\textbf{0.587} \\ ($\pm$ 0.010)} \\
      \specialrule{0.7pt}{0pt}{-1pt}
      \CodeAfter
      \begin{tikzpicture}
          \begin{scope}[shorten <=2pt, shorten >=2pt]
            \draw (1-|2) -- (4-|2);
            \draw (1-|5) -- (4-|5);
            \draw (1-|9) -- (4-|9);
        \end{scope}
      \begin{scope}[shorten <=2pt, shorten >=2pt]
            \draw (4-|2) -- (10-|2);
            \draw (4-|5) -- (10-|5);
            \draw (4-|9) -- (10-|9);
        \end{scope}
        \begin{scope}[shorten <=3pt, shorten >=3pt]
            \draw (2-|2) -- (2-|5);   
            \draw (2-|5) -- (2-|9);   
            \draw (2-|9) -- (2-|11);  
            \draw (3-|2) -- (3-|3);
            \draw (3-|3) -- (3-|5);
            \draw (3-|5) -- (3-|9);
            \draw (3-|9) -- (3-|11);
        \end{scope}
      \end{tikzpicture}
    \end{NiceTabular}%
    }
  \label{tab:appendix_ct_downstream_stat_sig}
\end{minipage}

%% file: checklist.tex
\section*{NeurIPS Paper Checklist}

\begin{enumerate}

\item {\bf Claims}
    \item[] Question: Do the main claims made in the abstract and introduction accurately reflect the paper's contributions and scope?
    \item[] Answer: \answerYes{}
    \item[] Justification: The abstract and introduction accurately reflect the paper's scope by highlighting \ours\ and its two core components, Sparsely Gated Alignment (SGA) for sparse vision-language correspondence and Dense Feature Regularization (DFR) for preserving fine-grained SSL features during alignment training. They also clearly describe \ours's zero-shot capability across classification, grounding, and segmentation from free-text queries on both 2D CXR and 3D CT, including the first zero-shot free-text segmentation on 3D CT without mask supervision.

\item {\bf Limitations}
    \item[] Question: Does the paper discuss the limitations of the work performed by the authors?
    \item[] Answer: \answerYes{}
    \item[] Justification: The paper explicitly discusses \ours's limitations across multiple sections including Section~\ref{sec:conclusion} (Conclusion): the modality and anatomy scope is restricted to chest X-ray and chest CT, and direct clinician assessment of clinical utility is left as future work (Section~\ref{sec:conclusion}); CT report generation shows a surface-level fluency trade-off, where METEOR and ROUGE-L trail the best baseline despite \ours-CT leading on clinical efficacy metrics (Section~\ref{sec:main_results}); and the ablation analysis identifies a sparsity--Dice trade-off when applying DFR alone, addressed by combining it with the gate confidence shift (Section~\ref{sec:sparse_precise}).

\item {\bf Theory assumptions and proofs}
    \item[] Question: For each theoretical result, does the paper provide the full set of assumptions and a complete (and correct) proof?
    \item[] Answer: \answerNA{}
    \item[] Justification: The paper does not include theoretical results; all contributions are empirical.

    \item {\bf Experimental result reproducibility}
    \item[] Question: Does the paper fully disclose all the information needed to reproduce the main experimental results of the paper to the extent that it affects the main claims and/or conclusions of the paper (regardless of whether the code and data are provided or not)?
    \item[] Answer: \answerYes{}
    \item[] Justification: The architecture is described in Section~\ref{sec:methods}; training data and protocol in Section~\ref{sec:experiments}; evaluation splits and class counts in Table~\ref{tab:appendix_test_dataset} (Appendix~\ref{apx:test_dataset}); hyperparameters, schedule, and downstream-head details in Appendix~\ref{apx:implementation_detail} with Table~\ref{tab:appendix_training_details}.

\item {\bf Open access to data and code}
    \item[] Question: Does the paper provide open access to the data and code, with sufficient instructions to faithfully reproduce the main experimental results, as described in supplemental material?
    \item[] Answer: \answerYes{}
    \item[] Justification: Our experiments are conducted entirely on publicly accessible datasets; details of the training and test data are described in Sections~\ref{sec:training_datasets} (training data) and \ref{sec:evaluation_datasets} (test data), with additional dataset details in Appendix~\ref{apx:test_dataset}, and baseline implementations follow the official code of existing works in their respective public repositories. We release our code at \url{https://github.com/deepnoid-ai/GLINT} and the model weights at \url{https://huggingface.co/Deepnoid/GLINT-CXR} and \url{https://huggingface.co/Deepnoid/GLINT-CT}, sufficient to reproduce the main experimental results.

\item {\bf Experimental setting/details}
    \item[] Question: Does the paper specify all the training and test details (e.g., data splits, hyperparameters, how they were chosen, type of optimizer) necessary to understand the results?
    \item[] Answer: \answerYes{}
    \item[] Justification: Training and downstream hyperparameters (batch size, learning rate, epochs, early-stopping patience, loss) are listed in Table~\ref{tab:appendix_training_details}; data splits in Table~\ref{tab:appendix_test_dataset}; preprocessing and downstream-head architectures are described in Appendix~\ref{apx:implementation_detail}.

\item {\bf Experiment statistical significance}
    \item[] Question: Does the paper report error bars suitably and correctly defined or other appropriate information about the statistical significance of the experiments?
    \item[] Answer: \answerYes{}
    \item[] Justification: Mean and 1-sigma standard deviation over three independent runs with different random seeds are reported in Tables~\ref{tab:appendix_cxr_zeroshot_stat_sig}--\ref{tab:appendix_ct_downstream_stat_sig} for \ours\ in zero-shot evaluation and for all evaluated methods in downstream evaluation; the variability source is the random seed, computed empirically across the three runs.

\item {\bf Experiments compute resources}
    \item[] Question: For each experiment, does the paper provide sufficient information on the computer resources (type of compute workers, memory, time of execution) needed to reproduce the experiments?
    \item[] Answer: \answerYes{}
    \item[] Justification: Compute resources for pretraining are reported in Table~\ref{tab:compute_cost} (Appendix~\ref{apx:implementation_detail}).

\item {\bf Code of ethics}
    \item[] Question: Does the research conducted in the paper conform, in every respect, with the NeurIPS Code of Ethics \url{https://neurips.cc/public/EthicsGuidelines}?
    \item[] Answer: \answerYes{}
    \item[] Justification: We have reviewed the NeurIPS Code of Ethics. Our work uses publicly available, de-identified medical imaging datasets (cited in Section~\ref{sec:experiments} and Appendix~\ref{apx:test_dataset}) under their respective licenses, conducts no new human-subjects research, and complies with each dataset's terms of use.

\item {\bf Broader impacts}
    \item[] Question: Does the paper discuss both potential positive societal impacts and negative societal impacts of the work performed?
    \item[] Answer: \answerYes{}
    \item[] Justification: Potential positive societal impacts of \ours, including reduced annotation cost, learning from long-tail disease descriptions, mitigation of inter-annotator disagreement in pixel-level supervision, and enabling language-centric clinical applications such as report drafting, are discussed in Section~\ref{sec:introduction}; deployment caveats, including the need for direct clinician assessment of clinical utility before any practical use, are discussed under the limitations paragraph in Section~\ref{sec:conclusion}.

\item {\bf Safeguards}
    \item[] Question: Does the paper describe safeguards that have been put in place for responsible release of data or models that have a high risk for misuse (e.g., pre-trained language models, image generators, or scraped datasets)?
    \item[] Answer: \answerNA{}
    \item[] Justification: The released artifacts are non-generative vision-language alignment encoders for radiology and do not fit the high-misuse-risk categories specified in the question (pre-trained language models, image generators, or scraped datasets); the training data are properly licensed medical imaging datasets, and the model produces no generative output.

\item {\bf Licenses for existing assets}
    \item[] Question: Are the creators or original owners of assets (e.g., code, data, models), used in the paper, properly credited and are the license and terms of use explicitly mentioned and properly respected?
    \item[] Answer: \answerYes{}
    \item[] Justification: All datasets and pretrained models used in this work are properly credited via citations and described in Sections~\ref{sec:training_datasets}, \ref{sec:evaluation_datasets}, and \ref{sec:implementation_details}; each is used in accordance with its respective license.

\item {\bf New assets}
    \item[] Question: Are new assets introduced in the paper well documented and is the documentation provided alongside the assets?
    \item[] Answer: \answerYes{}
    \item[] Justification: The code is released at \url{https://github.com/deepnoid-ai/GLINT} with a README describing setup and usage, and the \ours-CXR and \ours-CT model weights are released at \url{https://huggingface.co/Deepnoid/GLINT-CXR} and \url{https://huggingface.co/Deepnoid/GLINT-CT}.

\item {\bf Crowdsourcing and research with human subjects}
    \item[] Question: For crowdsourcing experiments and research with human subjects, does the paper include the full text of instructions given to participants and screenshots, if applicable, as well as details about compensation (if any)?
    \item[] Answer: \answerNA{}
    \item[] Justification: This work uses retrospective de-identified public radiology datasets only and does not involve crowdsourcing or new research with human subjects.

\item {\bf Institutional review board (IRB) approvals or equivalent for research with human subjects}
    \item[] Question: Does the paper describe potential risks incurred by study participants, whether such risks were disclosed to the subjects, and whether Institutional Review Board (IRB) approvals (or an equivalent approval/review based on the requirements of your country or institution) were obtained?
    \item[] Answer: \answerNA{}
    \item[] Justification: This work uses retrospective de-identified public radiology datasets and does not involve new prospective human-subjects research; the original datasets were collected and de-identified under their respective institutional protocols, as described in their original publications.

\item {\bf Declaration of LLM usage}
    \item[] Question: Does the paper describe the usage of LLMs if it is an important, original, or non-standard component of the core methods in this research? Note that if the LLM is used only for writing, editing, or formatting purposes and does \emph{not} impact the core methodology, scientific rigor, or originality of the research, declaration is not required.
    \item[] Answer: \answerYes{}
    \item[] Justification: We use \texttt{gpt-oss-120b}~\cite{agarwal2025gpt} for sentence-level decomposition of training reports (Section~\ref{sec:text_supervision}); the resulting sentences serve as alignment supervision for both \ours-CXR and \ours-CT, making this a non-standard, core-methodology use. The prompt design is described in Appendix~\ref{sec:appendix_prompts}.

\end{enumerate}